\documentclass[sigconf,nonacm]{acmart}

\usepackage{caption}
\usepackage[table]{xcolor}

\definecolor{blueA}{RGB}{198,219,239}  
\definecolor{blueB}{RGB}{158,202,225}  
\definecolor{blueC}{RGB}{107,174,214}  
\definecolor{blueD}{RGB}{49,130,189}   
\definecolor{blueE}{RGB}{8,48,107}     

\definecolor{redE}{RGB}{103,0,13}      
\definecolor{greenE}{RGB}{0,68,27}     
\definecolor{orangeE}{RGB}{127,39,4}   
\definecolor{maroonE}{RGB}{128,0,0}    

\usepackage{graphicx} 
\usepackage[inline]{enumitem}

\usepackage{tikz}
\usetikzlibrary{calc}
\usetikzlibrary{positioning, arrows.meta, shapes, fit, calc}

\usepackage{amsmath}
\usepackage{ragged2e}

\usepackage{minted}
\usepackage{multirow}
\usepackage{tabularx}

\usepackage{subcaption}

\usepackage{pgfplots}
\pgfplotsset{compat=1.18}
\graphicspath{{./images/}}

\usepackage{listings}
\usepackage{listings}

\usepackage{mdframed}
\usepackage{placeins}

\usepackage{filecontents}
\usepackage{booktabs}

\usepackage{tcolorbox}
\tcbuselibrary{skins,breakable}

\usepackage{xspace}
\newcommand{\BfPara}[1]{\vspace{0.3em}\noindent{\bf#1.}\xspace\xspace}
\newcommand{\ours}{\textsc{PromptAudit}\xspace}
\newcommand{\safe}{{\texttt{\textcolor{green!0!black}{SAFE}}}\xspace}
\newcommand{\vul}{{ \texttt{\textcolor{red!0!black}{VULNERABLE}}}\xspace}
\AtBeginDocument{%
  }

\usepackage{pifont}
\usepackage{amsmath}
\usepackage{graphicx}
\usepackage{booktabs}
\usepackage{multirow}
\usepackage{tabularx}
\usepackage{colortbl}
\usepackage{microtype}
\usepackage{balance}

\definecolor{f1strong}{HTML}{2D6A4F}
\definecolor{f1mod}   {HTML}{B7E4C7}
\definecolor{f1weak}  {HTML}{EF476F}
\definecolor{abshi}   {HTML}{EF476F}
\definecolor{abslo}   {HTML}{B7E4C7}
\definecolor{hdr}     {HTML}{2E4057}
\definecolor{altrow}  {HTML}{F5F7FA}

\usepackage{twemojis}

\newcommand{\xxmark}{\textcolor{black}{$\medcirc$}}
\newcommand{\yymark}{\textcolor{black}{$\medbullet$}}

\newcommand{\PA}{\textsc{PromptAudit}}

\newtcolorbox{rqbox}{
  enhanced,
  breakable,
  colback=blue!6,
  colframe=blue!15,
  boxrule=0.5pt,
  arc=3pt,
  left=5pt, right=5pt, top=4pt, bottom=4pt,
  notitle,
}

\setcopyright{none}
\renewcommand\footnotetextcopyrightpermission[1]{}
\begin{document}

\title{PromptAudit: Auditing Prompt Sensitivity in LLM-Based Vulnerability Detection}

\author{Steffen J. Camarato}
\affiliation{
  \institution{University of Central Florida}
  \city{Orlando}
  \state{Florida}
  \country{USA}
}
\email{sjcamarato@ucf.edu}

\author{Yahya Hmaiti}
\affiliation{
  \institution{University of Central Florida}
  \city{Orlando}
  \state{Florida}
  \country{USA}
}
\email{yohan.hmaiti@ucf.edu}

\author{Mandana Ghadamian}
\affiliation{
  \institution{University of Central Florida}
  \city{Orlando}
  \state{Florida}
  \country{USA}
}
\email{mandana.ghadamian@ucf.edu}

\author{David Mohaisen}
\affiliation{
  \institution{University of Central Florida}
  \city{Orlando}
  \state{Florida}
  \country{USA}
}
\email{mohaisen@ucf.edu}





\begin{abstract}
Large language models are increasingly used for vulnerability detection, yet their reliability under different prompt formulations remains uncharacterized. We present \ours, a controlled evaluation framework that isolates prompt effects by fixing the dataset, decoding, and parsing while varying only the prompting strategy. Using five prompting strategies across five open-weight models on 1,000 CVEs (6,074 code samples spanning 16 programming languages), we evaluate accuracy, recall, abstention, coverage, and effective F1. We find that standard chain-of-thought prompting achieves the strongest overall operational performance, while few-shot prompting provides model-dependent benefits that are most pronounced for prompt-sensitive models. In contrast, adaptive chain-of-thought frequently suppresses recall and self-consistency induces excessive abstention, sharply reducing effective performance. These results show that vulnerability detection behavior is jointly determined by the model and the prompt, and that prompt sensitivity is a first-class system property that must be explicitly characterized in evaluation and deployment.
\end{abstract}




\maketitle

\pagestyle{plain}   
\thispagestyle{plain}
\fancyhead{}        
\fancyfoot{}        
\renewcommand{\headrulewidth}{0pt}
\renewcommand{\footrulewidth}{0pt}

\section{Introduction}
Large language models (LLMs) are appearing more frequently in software-security tasks, including vulnerability detection, vulnerability repair, and security-aware code generation~\cite{lin_large_2025,jiang_investigating_2025,basic_vulnerabilities_2024, kharma2026security}. In this work, we focus specifically on general-purpose, instruction-tuned models applied to vulnerability detection through prompting alone, without any security-specific fine-tuning. They can read code, identify familiar weakness patterns, and provide rapid judgments that often appear more flexible than traditional static or dynamic analysis tools. Unlike traditional static and dynamic analysis tools based on fixed rules and signatures, LLMs operate over learned representations that generalize across programming languages and vulnerability classes~\cite{he_large_2023,ullah2024llms}. However, prior work also consistently reports that LLM-based vulnerability detection exhibits substantial variability, with performance strongly dependent on prompt formulation, model configuration, dataset properties, and evaluation protocol~\cite{sclar_quantifying_2024,chatterjee_posix_2024,lin_large_2025}.

Existing evaluations typically attribute this variability to model behavior alone. Yet, as recent analyses indicate, prompt sensitivity reflects not only model uncertainty but also how evaluation design mediates inference~\cite{sclar_quantifying_2024,liang_holistic_2023}. Semantically equivalent prompts can yield different reasoning traces and different vulnerability labels, indicating that reported performance depends on prompt structure, output constraints, and parsing rules in addition to the model itself~\cite{chatterjee_posix_2024,lin_large_2025}. In binary classification settings such as labeling code as \safe or \vul, this instability complicates reproducibility and obscures comparison across studies. For a simple binary task such as deciding whether a code snippet is \safe or \vul, this instability carries security-specific consequences that go beyond generic classification error. Vulnerability detection is not a symmetric problem: a false negative, which is a vulnerable snippet classified as safe, may allow exploitable code to proceed undetected, whereas a false positive adds reviewer workload but does not introduce risk. Abstention, where the model declines to produce a verdict, poses a similar operational concern: code that receives no verdict is effectively unreviewed unless a downstream fallback is in place. Prompt-induced shifts in recall or abstention rate can therefore alter the operational security behavior of a detector without any change to the underlying model, making prompt design a first-order deployment concern rather than an evaluation detail.

Related work further shows that evaluation artifacts can amplify apparent instability. Scoring methods, prompt templates, and output extraction procedures may penalize semantically correct outputs or shift decision thresholds without improving underlying reasoning~\cite{huaFlawArtifactRethinking}. In vulnerability detection, these effects are compounded by dataset characteristics, including class imbalance and commit-level labeling, where ground truth is often ambiguous or context-dependent~\cite{lin_large_2025,ullah2024llms}. As a result, observed accuracy may conflate model capability, prompt-induced variance, and dataset uncertainty, limiting the interpretability of empirical conclusions~\cite{liang_holistic_2023}.

This work reframes prompt sensitivity as an evaluation problem rather than a property of individual prompts or models. Building on prior analyses of prompting strategies, sensitivity, and benchmarking practices, we introduce \ours, a controlled evaluation framework that isolates the effect of prompt structure by fixing the dataset, decoding parameters, and evaluation pipeline while varying only the prompting strategy. This study does not propose a new vulnerability detection model, introduce new benchmarks, or claim improvements in detection accuracy. Instead, it provides a systematic measurement analysis of how inference-time prompting strategies affect the stability and reliability of LLM-based vulnerability classification. By disentangling prompt-induced variance from other experimental factors, \ours{} enables more reliable interpretation of prior results and supports principled comparison across vulnerability detection studies.

\BfPara{Research Questions} This study addresses four research questions. \textbf{RQ1:} How do inference-time prompting strategies affect vulnerability detection performance, coverage, and abstention under controlled evaluation conditions? \textbf{RQ2:} Do prompting strategies primarily improve conditional detection performance, or do they shift the precision-recall-coverage tradeoff? \textbf{RQ3:} What recurring failure modes emerge under different prompting strategies, and how do they affect operational reliability? \textbf{RQ4:} How do model-strategy interactions shape operational performance, and which combinations provide the most reliable behavior under our evaluation setup? Sections~\ref{sec:evaluation-results} and~\ref{sec:discussion} address these questions.

\BfPara{Contributions} We make the following contributions:
\begin{enumerate*}
  \item {\bf Prompt Sensitivity Controlled Evaluation.}
  We present a controlled measurement study of prompt sensitivity in LLM vulnerability detection that isolates prompting strategy as the primary experimental variable. By fixing the dataset, decoding parameters, and evaluation pipeline, we disentangle prompt-induced variance from model configuration and evaluation artifacts.

  \item {\bf Systematic Analysis of Inference-Time Prompting Strategies.}
  We compare five concrete prompt templates corresponding to zero-shot, few-shot, chain-of-thought, adaptive chain-of-thought, and self-consistency under a unified setup, showing how prompt-template choice affects vulnerability classification across models.

  \item {\bf Coverage-Aware Analysis of Operational Reliability.}
  We show that prompting strategies affect not only standard F1, but also abstention, coverage, and effective F1, revealing failure modes such as recall collapse and coverage collapse that are obscured by accuracy-only evaluation.

  \item {\bf \textsc{PromptAudit}: A Reproducible Evaluation Framework.}
  We introduce \ours, a reusable evaluation framework for auditing prompt-induced variance in LLM-based vulnerability detection. \ours enables reproducible measurement studies and principled comparison across prompting strategies without introducing new detection models or benchmarks.
\end{enumerate*}

\BfPara{Scope of Claims}
This study makes comparative measurement claims rather than claims about absolute vulnerability-detection capability. We evaluate how fixed prompt-template instances change model behavior under one CVEfixes-derived label source, one decoding configuration, and one default parser policy. The labels provide a shared experimental substrate, but do not establish that every isolated snippet is independently exploitable or safe. Similarly, each prompting strategy is represented by one concrete template instance; results should therefore be interpreted as template-level evidence rather than universal claims about all possible zero-shot, few-shot, CoT, adaptive CoT, or self-consistency prompts.

\section{Related Work}
This section situates our work within prior work on learning-based vulnerability detection, inference-time prompting strategies, and LLMs evaluation robustness. Rather than exhaustively surveying all related systems, we focus on work most relevant to understanding how prompting, dataset construction, and evaluation design interact to shape reported performance. This perspective clarifies the limitations of existing evaluations and motivates our controlled analysis of prompt-induced variance in LLM-based detection.

\subsection{Vulnerability Detection and Prompting}

The prior work on automated vulnerability detection spans static and dynamic analysis tools~\cite{chess2007secure,viega2000its4,serebryany2012addresssanitizer}, learning-based approaches operating over token sequences and program graphs~\cite{li_vuldeepecker_2018,russell2018automated,zhou2019devign,chakraborty2021deep}, and pre-trained code models such as CodeBERT and GraphCodeBERT~\cite{feng2020codebert,guo2020graphcodebert}. Security-oriented variants, including VulBERTa and LineVul, demonstrate that large-scale pretraining yields representations transferable to vulnerability detection~\cite{hanif2022vulberta,fu_linevul_2022}.

Building on these advances, instruction-tuned LLMs extend this line of work by enabling vulnerability classification through natural language prompts without task-specific fine-tuning. Empirical studies on GPT-3.5, GPT-4~\cite{pearce2023examining,siddiq2022empirical}, and code models such as Code Llama, StarCoder, and DeepSeek-Coder~\cite{roziere2023code,li2023starcoder,guo2024deepseek} show that LLMs can recognize vulnerability patterns, but with performance highly dependent on vulnerability type, code complexity, and prompting strategy. Recent comparative evaluations further demonstrate substantial variation across models, languages, and prompt settings, with some configurations performing near chance~\cite{lin_large_2025,jiang_investigating_2025}.

A recurring limitation of LLM-based detection is high sensitivity to prompt formulation and inference parameters~\cite{white2024chatgpt,liu2024lost,wei2022chain,sclar_quantifying_2024}. Unlike traditional analyzers with fixed decision procedures~\cite{chess2007secure,johnson2013don}, LLMs perform probabilistic inference over multi-step reasoning paths, making predictions sensitive to instruction framing, example selection, and output constraints. While techniques such as self-consistency can reduce variance across generations~\cite{wang_self-consistency_2023}, prompt-induced instability remains a major obstacle to reliable evaluation.

Despite this growing interest, existing studies largely focus on proprietary models, limiting systematic comparison of open-weight alternatives~\cite{fan2023large}. Moreover, while prompt engineering is known to affect LLM behavior~\cite{wei2022chain,kojima2022large}, most evaluations consider a narrow set of prompt variants or conflate prompt effects with model, dataset, or decoding choices. As a result, reported improvements often obscure whether gains reflect improved reasoning or prompt-induced shifts on specific benchmarks, such as CVEfixes and Big-Vul~\cite{bhandari_cvefixes_2021,fan_cc_2020}.

\subsection{Prompt Sensitivity and Evaluation Design}

To better understand these limitations, recent work directly quantifies prompt sensitivity. Sclar et al.~\cite{sclar_quantifying_2024} show that prompt \emph{format} alone can induce accuracy swings of up to 76\%, and that neither scaling nor instruction tuning reliably mitigates this variance. Chatterjee et al.~\cite{chatterjee_posix_2024} introduce POSIX, a sensitivity index based on log-likelihood changes under intent-preserving substitutions, and show that sensitivity is structured rather than random: template-level changes dominate for multiple-choice tasks, while paraphrasing has stronger effects on open-ended generation. Beyond quantification, other work focuses on mitigation strategies, including diversifying prompt formats~\cite{ngweta_towards_2025} and adopting semantics-aware evaluation methods to reduce artifacts introduced by rigid scoring and parsing rules~\cite{huaFlawArtifactRethinking}. These studies suggest that prompt sensitivity is partly entangled with the evaluation harness itself rather than reflecting model behavior alone.

In the specific context of vulnerability detection, another line of work shows that prompt sensitivity is further compounded by dataset construction. Common benchmarks derive labels from vulnerability-fixing commits linked to CVE records~\cite{bhandari_cvefixes_2021,fan_cc_2020,chen_diversevul_2023}. This construction supports large-scale studies, but it can mislabel isolated snippets when exploitability depends on calling context, configuration, or interprocedural data flow~\cite{chakraborty2021deep,arp2022and,steenhoek_empirical_2023}. Recent LLM evaluations show that these label and context limitations interact with prompt choice, language, and evaluation protocol, with performance often degrading under stricter task definitions~\cite{lin_large_2025,ullah2024llms,jiang_investigating_2025}.

\begin{table}[t]
\centering
\footnotesize
\caption{Comparison of prior work along evaluation design dimensions. C1--C6 denote: use of LLMs, prompt variation, dataset control, decoding control, evaluation control, and prompt isolation. Prior work typically varies prompts alongside other factors, obscuring attribution, whereas our design isolates prompt effects under controlled conditions.}\label{tab:related_work_comparison}\vspace{-2mm}
\begin{tabular}{lcccccc}
\hline
\textbf{Work} & \textbf{C1} & \textbf{C2} & \textbf{C3} & \textbf{C4} & \textbf{C5} & \textbf{C6} \\
\hline
Pearce et al.~\cite{pearce2023examining}        & \yymark & \xxmark & \xxmark & \xxmark & \xxmark & \xxmark \\
Siddiq et al.~\cite{siddiq2022empirical}        & \yymark & \xxmark & \xxmark & \xxmark & \xxmark & \xxmark \\
Lin et al.~\cite{lin_large_2025}                & \yymark & \yymark & \xxmark & \xxmark & \xxmark & \xxmark \\
Jiang et al.~\cite{jiang_investigating_2025}    & \yymark & \yymark & \xxmark & \xxmark & \xxmark & \xxmark \\
Sclar et al.~\cite{sclar_quantifying_2024}      & \yymark & \yymark & \xxmark & \xxmark & \xxmark & \xxmark \\
Chatterjee et al.~\cite{chatterjee_posix_2024}  & \yymark & \yymark & \xxmark & \xxmark & \xxmark & \xxmark \\
Ngweta et al.~\cite{ngweta_towards_2025}        & \yymark & \yymark & \xxmark & \xxmark & \xxmark & \xxmark \\
Hua et al.~\cite{huaFlawArtifactRethinking}     & \yymark & \yymark & \xxmark & \xxmark & \xxmark & \xxmark \\
\hline
\textbf{\ours}                              & \yymark & \yymark & \yymark & \yymark & \yymark & \yymark \\
\hline
\end{tabular}
\end{table}

Overall, the prior work shows that prompt sensitivity, dataset ambiguity, and evaluation design jointly shape reported LLM vulnerability detection performance. Existing studies typically vary these factors together or treat prompts as fixed inputs, as summarized in Table~\ref{tab:related_work_comparison} and compared with our work, making it difficult to attribute observed performance differences to any single source. This work, on the other hand, addresses that gap by isolating prompt structure as a variable while fixing the dataset, inference parameters, and evaluation pipeline, enabling a controlled analysis of how prompting alone influences detection outcomes.

\section{Methodology}
\label{sec:methodology}

\begin{figure}
    \centering
\begin{tikzpicture}[
    font=\footnotesize,
    node distance=0.1cm and 0.2cm,
    box/.style={draw, rounded corners, thick, align=center, minimum width=1.0cm, minimum height=0.6cm},
    greenbox/.style={box, draw=green!60!black, fill=green!10, thick},
    topbox/.style={box, draw=blue!70!black, fill=blue!10, thick},
    outbox/.style={box, draw=orange!70!black, fill=orange!15, thick},
    arrow/.style={-Latex, thick}
]

\node[topbox] (dataset)  at (0,-1) {{\bf Dataset}$^\star$};
\node[topbox, right=.5cm of dataset] (decode) {{\bf Decoding Configuration}$^\star$};
\node[topbox, right=0.5cm of decode] (parsercfg) {{\bf Parser Configuration}$^\star$};

\node[box, below=0.5cm of dataset] (loader)
{{\bf Loader}\\
\footnotesize Samples and \\ true labels};

\node[greenbox, right=of loader] (prompt)
{{\bf Prompt}\\
\footnotesize One of five \\ strategies};

\node[box, right=of prompt] (model)
{{\bf Execution}\\
\footnotesize LLM generates\\ output};

\node[box, right=of model] (parser)
{{\bf Parser}\\
\footnotesize Extract \\verdict};

\node[box, right=of parser] (metrics)
{{\bf Metrics Engine}\\
\footnotesize (Acc, Recall, \\Abstention)};

\draw[arrow] (loader) -- (prompt);
\draw[arrow] (prompt) -- (model);
\draw[arrow] (model) -- (parser);
\draw[arrow] (parser) -- (metrics);

\node[draw, dashed, thick, rounded corners,
      fit=(loader)(prompt)(model)(parser)(metrics),
      inner sep=2pt] (cluster) {};

\node[above=-0.04cm of prompt, text=green!60!black]
{\textbf{VARYING COMPONENT}};

\node[circle, fill=black, inner sep=1.0pt] (b1) at ($(cluster.north west)!0.098!(cluster.north east)$) {};
\node[circle, fill=black, inner sep=1.0pt] (b2) at ($(cluster.north west)!0.41!(cluster.north east)$) {};
\node[circle, fill=black, inner sep=1.0pt] (b3) at ($(cluster.north west)!0.813!(cluster.north east)$) {};

\draw[arrow] (dataset.south) -- (b1);
\draw[arrow] (decode.south) -- (b2);
\draw[arrow] (parsercfg.south) -- (b3);

\node[outbox, below=0.6cm of metrics] (output)
{Output};

\draw[arrow] (metrics) -- (output);

\node[below=0.0cm of cluster, xshift=0cm] (cart)
{\textbf{Cartesian Product of Experimental Factors}};

\node[box, below left=0.05cm and 0.25cm of cart] (D) {\textbf{D} (1)};
\node[box, right=0.25cm of D] (M) {\textbf{M} (5)};
\node[box, right=0.25cm of M] (P) {\textbf{P} (5)};
\node[box, right=0.25cm of P] (O) {\textbf{O} (3)};
\node[box, right=0.25cm of O] (R) {\textbf{R} (2)};

\node at ($(D)!0.5!(M)$) {$\times$};
\node at ($(M)!0.5!(P)$) {$\times$};
\node at ($(P)!0.5!(O)$) {$\times$};
\node at ($(O)!0.5!(R)$) {$\times$};
\end{tikzpicture}\vspace{-2mm}
        \caption{\ours evaluation pipeline. The main evaluation fixes dataset, decoding, output protocol, and parser mode while varying the prompt template and is evaluated against dataset (D), models (M), prompts (P), outputs (O), and reranking (R). $^\star$ indicates a fixed component.}
        \label{fig:pipeline}\vspace{-3mm}
\end{figure}
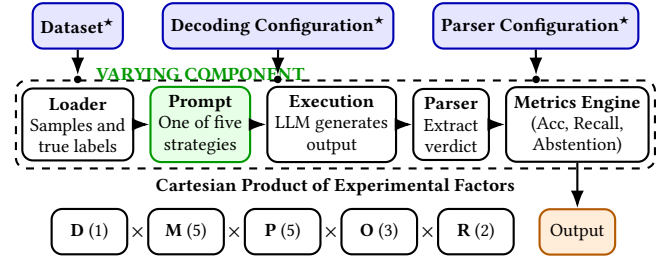

We review the \ours framework, including the evaluation datasets, prompting strategies, and pipeline architecture. We describe the system's operational mechanics and the theoretical rationale underpinning each design specification. Figure~\ref{fig:pipeline} summarizes the evaluation pipeline, while Figure~\ref{fig:strategy-spectrum} summarizes the prompt-design spectrum. We describe the \ours framework, including the evaluation datasets, prompting strategies, and pipeline architecture.

\subsection{System Overview}

\ours is a modular evaluation framework that keeps \textit{dataset loading}, \textit{prompt formatting}, \textit{model execution}, \textit{output parsing}, and \textit{metric computation} separate. This structure keeps the experimental substrate fixed while the prompt-template instance changes, reducing the chance that observed variation is caused by execution or reporting differences rather than prompt structure. At a high level, the runner loads a labeled code sample, formats it with the selected prompt template and output protocol, queries an open-weight model through the Ollama backend, parses the response into \safe, \vul, or \texttt{UNKNOWN}, and computes accuracy, precision, recall, abstention, coverage, F1, and effective F1. Parser-mode and output-protocol ablations use the same pipeline to test whether extraction or verdict-placement choices materially affect the measured prompt sensitivity.

\subsection{Dataset Construction}

To evaluate vulnerability detection under conditions aligned with real-world practice, we derive our dataset from CVEfixes~\cite{bhandari_cvefixes_2021}, a public repository linking CVE records to vulnerability-fixing commits in open-source projects. CVEfixes contains {\bf 9,386 CVE records} with {\bf 75,950 code samples} across multiple languages. We randomly sampled {\bf 1,000 CVEs} from the full corpus, with the sample proportionally reflecting the year-of-disclosure distribution present in CVEfixes across 1999--2024. This provides natural temporal coverage without overrepresenting any single disclosure period, toolchain, or coding convention.
This design prioritizes breadth and reproducibility rather than completeness. Each sampled CVE maps to multiple code samples because CVEfixes links each vulnerability to all modified files in the fixing commit. Hence, this yielded {\bf 6,074 code samples} for evaluation from 1,000 CVEs after filtering duplicates and non-code artifacts. The 1,000-CVE subset size was validated empirically: we evaluated subsets at 50, 100, 500, and 1,000 CVE levels and found that performance trends across prompting strategies stabilized at 1,000 CVEs, representing the smallest tractable configuration for controlled, repeatable experimentation. The curated subset will be released publicly alongside the artifacts.

The resulting 6,074 samples span 16 programming languages, with no single language exceeding 26.54\% of the corpus, reflecting the natural distribution of CVEfixes rather than deliberate language-level balancing. The samples are balanced by construction: each CVE contributes both a pre-fix (\vul) and a post-fix (\safe) version from the same commit, yielding a near-equal class distribution after deduplication and filtering. All samples are drawn from standardized CVE documentation and linked open-source repositories, preserving alignment with public security disclosures while supporting controlled evaluation. As with other commit- and disclosure-derived datasets, the presence of a CVE does not imply that every associated code snippet independently exhibits exploitable behavior in isolation.

Each dataset entry includes the CVE identifier, disclosure year, programming language, and vulnerability metadata extracted from the corresponding CVE record. Entries further include one or more associated code snippets reflecting the reported vulnerability and its surrounding context, as available in the referenced repository. This structure supports consistent evaluation across samples while making explicit the limitations inherent in snippet-level representations of context-dependent vulnerabilities.

\BfPara{Data Cleaning and Limitations} Samples from CVEfixes were scrubbed before integration into the \ours pipeline to address data quality issues. We removed: (1) non-code artifacts (e.g., configuration files, documentation, README), (2) files failing programming language validation, (3) exact and near-duplicates ($> 98\%$ similarity), and (4) snippets $< 10$ lines (all manually verified). The runtime loader then retains only before/after file pairs whose original filename can be recovered from the CVE folder metadata and whose extension maps to a supported programming language; files with unknown language are excluded.

We note that CVEfixes exhibits limitations inherent to commit-based datasets. These issues appear as: (1) \textit{Labeling ambiguity}, where commits may include non-security changes alongside the fix, mitigated by retaining only files directly modified in the fixing commit and excluding non-code artifacts; (2) \textit{Context dependency}, where some snippets require broader program context to assess exploitability, acknowledged as an inherent constraint of snippet-level evaluation; and (3) \textit{Incomplete fixes}, where CVEs map to multi-stage patches, addressed by deduplication and near-duplicate filtering at the $>$98\% similarity threshold. Additionally, these label quality constraints apply uniformly across all evaluated prompting strategies and models. Any residual noise applies across all evaluated conditions and is therefore unlikely to fully explain the differential prompt effects observed in Section~\ref{sec:evaluation-results}, which are the primary claims of this work.

Note that we do not interpret CVEfixes labels as proof that each isolated snippet is independently exploitable. The labels provide a fixed experimental substrate for comparing prompt-induced variation under the same data source. Our claims therefore concern relative changes across prompt templates under a shared label construction procedure, not absolute vulnerability-detection accuracy.

\BfPara{Dataset Preprocessing}
Prior to prompting, all samples are normalized into a unified representation to ensure consistency across experimental conditions. Each sample is structured as a CVE-linked record containing a unique identifier, inferred programming language, raw code snippet, and a binary label (\safe or \vul). Retained samples were also standardized in terms of whitespace and indentation. This normalization is intended to standardize model inputs rather than to encode complete vulnerability semantics.

Vulnerable and fixed code samples are derived from standardized before/after file structures associated with vulnerability-fixing commits. Code appearing prior to the fix is labeled \vul, while the patched version is labeled \safe. Programming language is inferred from file extensions, and non-code artifacts (e.g., license or documentation) are excluded to reduce ambiguity during analysis.

We adopt a binary \safe/\vul as a controlled abstraction for studying prompt-induced variance under fixed conditions. This labeling scheme is not intended to capture the full semantics of real-world vulnerabilities, which may depend on execution context, configuration, or interprocedural behavior beyond what is represented in isolated code snippets. Moreover, when a prompt does not produce a \safe/\vul label, the output is treated as an abstention and recorded as \texttt{UNKNOWN} for evaluation purposes.

\subsection{Prompting Strategies}

\begin{figure}[t]
\centering
\resizebox{\columnwidth}{!}{
\begin{tikzpicture}[
    font=\footnotesize,
    panel/.style={
        draw,
        rounded corners,
        thick,
        align=center,
        minimum width=2.0cm,
        minimum height=3.7cm,
        text width=2.0cm,
        anchor=north,
        inner sep=4pt
    },
    innerbox/.style={
        draw,
        rounded corners,
        dotted,
        align=center,
        minimum width=1.85cm,
        minimum height=1.35cm,
        inner sep=3pt
    },
    bottomlabel/.style={
        align=center,
        anchor=north,
        text width=2.8cm
    },
    arrow/.style={-Latex, thick}
]

\node[panel, draw=black] at (0,0) (zs) {
\begin{minipage}[t][4.15cm][t]{2.0cm}
\centering
{\bf \ding{182} ZS}\\[3pt]

\begin{minipage}[t][0.78cm][t]{2.0cm}
\centering
{\it Model answers}\\[-1pt]
{\it with only the}\\[-1pt]
{\it task instruction.}
\end{minipage}\\[6pt]

\begin{tikzpicture}
\node[innerbox] {
Is the following\\
code vulnerable?\\
SAFE or\\
VULNERABLE
};
\end{tikzpicture}

\vfill
{\bf Instruction}
\end{minipage}
};

\node[panel, draw=black, right=0.2cm of zs] (fs) {
\begin{minipage}[t][4.15cm][t]{2.0cm}
\centering
{\bf \ding{183} FS}\\[3pt]

\begin{minipage}[t][0.78cm][t]{2.0cm}
\centering
{\it Model sees}\\[-1pt]
{\it a few examples.}
\end{minipage}\\[6pt]

\begin{tikzpicture}
\node[innerbox, fill=green!10] {
Code 1 $\rightarrow$ SAFE\\
Code 2 $\rightarrow$\\
VULNERABLE\\
$\dots$
};
\end{tikzpicture}

\vfill
{\bf Instruction}\\[-1pt]
{\bf + Examples}
\end{minipage}
};

\node[panel, draw=black, right=0.2cm of fs] (cot) {
\begin{minipage}[t][4.15cm][t]{2.0cm}
\centering
{\bf \ding{184} CoT}\\[3pt]

\begin{minipage}[t][0.78cm][t]{2.0cm}
\centering
{\it Model is asked}\\[-1pt]
{\it to reason}\\[-1pt]
{\it step-by-step}
\end{minipage}\\[6pt]

\begin{tikzpicture}
\node[innerbox, fill=orange!10] {
1. Understand\\
2. Identify\\
3. Analyze\\
4. Conclude\\
{\bf Answer: SAFE}\\
{\bf / VULNERABLE}
};
\end{tikzpicture}

\vfill
{\bf Instruction}\\[-1pt]
{\bf + CoT Cue}
\end{minipage}
};

\node[panel, draw=black, right=0.2cm of cot] (acot) {
\begin{minipage}[t][4.15cm][t]{2.0cm}
\centering
{\bf \ding{185} A-CoT}\\[3pt]

\begin{minipage}[t][0.78cm][t]{2.0cm}
\centering
{\it Adaptive reasoning}
\end{minipage}\\[6pt]

\begin{tikzpicture}
\node[innerbox, fill=purple!10] {
Assess complexity\\
$\downarrow$\\
More reasoning?\\
Yes $\rightarrow$ Continue\\
No $\rightarrow$ Answer
};
\end{tikzpicture}

\vfill
{\bf Instruction}\\[-1pt]
{\bf + A-CoT Cue}
\end{minipage}
};

\node[panel, draw=black, right=0.2cm of acot] (sc) {
\begin{minipage}[t][4.15cm][t]{2.0cm}
\centering
{\bf \ding{186} S-C}\\[3pt]

\begin{minipage}[t][0.78cm][t]{2.0cm}
\centering
{\it Multiple paths}\\[-1pt]
{\it + aggregation}
\end{minipage}\\[6pt]

\begin{tikzpicture}
\node[innerbox, fill=red!10] {
Path 1 $\rightarrow$ VULN\\
Path 2 $\rightarrow$ VULN\\
$\dots$\\
Path n $\rightarrow$ SAFE\\
{\bf Majority Vote}
};
\end{tikzpicture}

\vfill
{\bf Instruction}\\[-1pt]
{\bf + S-C Cue}
\end{minipage}
};

\draw[arrow]
([yshift=-0.8cm]zs.south west) --
([yshift=-0.8cm]sc.south east);

\node[bottomlabel] at ([yshift=-1.00cm]zs.south) {
Low Structure\\
Minimal Guidance
};

\node[bottomlabel] at ([yshift=-1.00cm]sc.south) {
High Structure\\
\makebox[0pt][c]{Strong Guidance \& Aggregation}
};

\end{tikzpicture}
}\vspace{-3mm}
\caption{Prompt strategy design spectrum. The five strategies, ZS, FS, CoT, A-CoT, and S-C, form a continuum from minimal guidance to structured, aggregated reasoning.}
\label{fig:strategy-spectrum}\vspace{-5mm}
\end{figure}
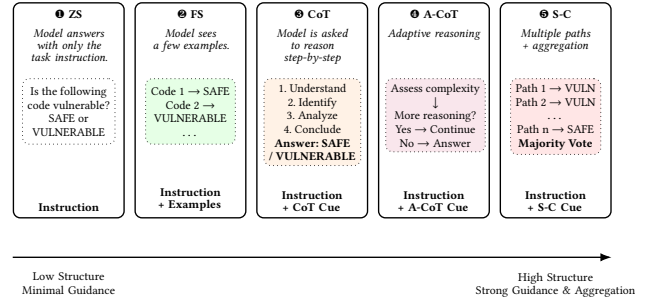

\ours evaluates five prompt templates corresponding to common prompting strategies drawn from practice and recent prompt-engineering research \cite{chen2025unleashing}. The templates vary in imposed structure, encouragement of reasoning, and model freedom in answer generation. Since prompting is the variable under study, each strategy is implemented as a configurable template via the prompt loader. Representative excerpts are included in Appendix~\ref{appendix:prompts-parser}. The strategies we used span a spectrum from minimal guidance to highly structured reasoning, enabling \ours to isolate how instruction level influences model behavior. Figure~\ref{fig:strategy-spectrum} summarizes this progression from minimal prompting to increasingly structured and aggregated reasoning.

\begin{itemize}[leftmargin=10pt, labelindent=10pt,itemsep=0mm]
    \item \textit{Zero-Shot (ZS).} The model receives only the task instruction and direct classification guidance. No examples or step-by-step reasoning are provided. This serves as the baseline condition, with exact verdict placement controlled separately by the selected output protocol.

    \item {\textit{Few-Shot (FS).}} A small set of labeled example classifications is provided before the target snippet. In the current implementation, these are synthetic CVE-style before/after function examples. Few-shot prompting is widely used because it can reduce misclassification in low-context tasks without explicitly requesting step-by-step reasoning.

    \item \textit{Chain-of-Thought (CoT).} The model is prompted to reason step by step about factors such as input validation, memory safety, race conditions, and injection or trust-boundary risks. The selected output protocol determines whether the decisive label appears on the first line or on the final line after the explanation.

    \item \textit{Adaptive Chain-of-Thought (A-CoT).} The model is instructed to reason only when needed. Trivially safe or vulnerable code receives minimal explanation, while complex cases (e.g., pointer arithmetic, raw memory manipulation, manual resource management, or complex input handling) elicit more detailed reasoning. As with CoT, the selected output protocol determines whether the final verdict is emitted first or last.

    \item \textit{Self-Consistency (S-C).} The model produces multiple independent predictions using the A-CoT template under the active output protocol. Each vote is parsed using the selected parser mode and aggregated by majority voting. Thus, the S-C results evaluate majority voting over A-CoT-style generations, not self-consistency over all possible base prompts. When no label receives a true majority across the requested samples, the final output is treated as an abstention (\texttt{UNKNOWN}).
\end{itemize}

\BfPara{Output-Protocol and Parser Ablations} Beyond prompt family, \ours exposes two controlled ablation axes. The output protocol toggles whether the model must emit the verdict on the first line (\texttt{verdict\_first}) or on the final line after any explanation (\texttt{verdict\_last}). The parser mode toggles whether label extraction is restricted to the exact protocol location (\texttt{strict}), broadened to explicit verdict phrases (\texttt{structured}), or further extended with lexical fallback (\texttt{full}). These ablations allow us to measure whether reported prompt effects depend on verdict placement or extraction heuristics rather than on prompt family alone. Unless otherwise stated, the prompt-strategy results reported in Section~\ref{sec:evaluation-results} correspond to the default \texttt{verdict\_first} + \texttt{full} configuration.

\noindent\textbf{Prompt Input Construction}
Prior to prompting, no semantic rewriting, truncation, or vulnerability-specific
annotation is applied to the code. Snippets are passed verbatim to preserve real-world
complexity and prevent cue leakage that could confound the experiment. Each prompt is constructed by inserting a single code sample into a configurable prompt template corresponding to the evaluated prompting strategy. The template specifies the instruction framing, output constraints, and optional reasoning structure (when the prompt strategy involves reasoning), while the underlying code content remains unchanged across conditions. This process allows observed
differences to be attributed to the prompting strategy rather than data transformation artifacts.

\subsection{Evaluation Pipeline Architecture}
The pipeline structure helps keep the evaluations consistent and repeatable across all models and prompting strategies. The pipeline begins with the dataset loader reading the dataset and preparing each snippet with its ground-truth label. The prompt loader constructs the strategy-specific prompt, after which the runner appends the active output-protocol suffix that governs verdict placement. The model loader initializes each open-weight architecture under identical decoding settings. By abstracting model invocation behind a common \texttt{generate()} method, \ours ensures model differences do not arise from backend inconsistencies.

The experiment runner cycles through every combination of dataset, model, prompting strategy, output protocol, parser mode, and code sample, corresponding to the Cartesian product $\mathcal{D} \times \mathcal{M} \times \mathcal{P} \times \mathcal{O} \times \mathcal{R}$. Per configuration, it builds the strategy prompt, appends the selected output-protocol instruction, sends the resulting prompt to the model, and records the model's raw output. Each response gets forwarded to the label parser, which applies the selected parser mode (Section~\ref{sec:label-parser}). Afterward, the output is assigned one of three labels: \safe, \vul, or \texttt{UNKNOWN}, with the last category used when the response is ambiguous, inconsistent, or cannot be mapped to a valid verdict under the chosen parser mode.

Parsed predictions are passed to the metrics engine, which computes accuracy, precision, recall, abstention rate, coverage, standard F1, and effective F1 for each model-prompt combination. The reporting module then aggregates results into CSV summaries and produces an interactive HTML dashboard that visualizes trends, model comparisons, and the effects of prompting.

\subsection{Model Settings and Backend Configuration}

\ours relies on five open-weight models served through the local Ollama backend:
Mistral (\texttt{mistral:latest}), Gemma 7B (\texttt{gemma:7b}),
CodeLlama 7B Instruct (\texttt{codellama:7b-instruct}),
Falcon 7B Instruct (\texttt{falcon:7b-instruct}), and DeepSeek-Coder 6.7B Instruct (\texttt{deepseek-coder:6.7b-instruct}). These models represent a mix of general-purpose and code-oriented architectures commonly used for local analysis tasks and size is constrained to roughly 7B parameters or smaller due to hardware limitations. To avoid introducing variation unrelated to prompting, all models run under identical
decoding parameters: temperature~0.2, top-p~0.9, top-k~40, a 250-token limit, and fixed
sampling penalties. These settings reduce sampling randomness and help prevent the
formatting drift that often occurs at higher temperatures. Appendix~\ref{appendix:gen-settings}
lists the full configuration. By holding decoding constant, \ours isolates prompt structure as the primary
source of output variation.

\subsection{Output-Protocol Ablations}
\label{sec:output-protocol}

Because LLMs often embed labels within explanations or defer decisions, \ours defines two output-protocol variants. Under \texttt{verdict\_first}, the model must place exactly one word, \safe or \vul, on the first non-empty line, with any explanation following. Under \texttt{verdict\_last}, the explanation precedes the verdict, and the final non-empty line must contain only \safe or \vul. This separation isolates formatting effects from prompt design: prompting controls reasoning, while the protocol controls label placement. Unless otherwise stated, results use the \texttt{verdict\_first} configuration. For tractability, the protocol ablation is evaluated on a stratified 1,232-sample subset of the 6,074-sample dataset, preserving language and prompt distributions. As shown in Section~\ref{sec:ablation}, \texttt{verdict\_first} yields stronger operational performance and is therefore used as the default.

\subsection{Label Parsing Logic}
\label{sec:label-parser}

Even with controlled output protocols, models occasionally place the label in the wrong location, phrase it in a longer sentence, or embed it in freer reasoning text. \ours therefore exposes three parser modes built on the same layered parser. \textit{Strict parser.} The parser checks only the protocol-defined verdict location: the first non-empty line for \texttt{verdict\_first} or the last non-empty line for \texttt{verdict\_last}. The target line must contain a single \safe/\vul token after minor normalization of punctuation. \textit{Structured parser.} If strict parsing fails, the parser scans non-empty lines from bottom to top for explicit verdict markers such as \texttt{Final answer: SAFE}, \texttt{The final answer is SAFE}, \texttt{In conclusion, the code is VULNERABLE}, or \texttt{The code is SAFE}. \textit{Full parser.} If neither of the above succeeds, the parser performs a whole-response lexical fallback over safety and vulnerability cues, including \texttt{unsafe}, \texttt{vulnerable}, \texttt{vulnerabilities}, \texttt{exploitable}, \texttt{at risk}, \texttt{safe}, \texttt{secure}, \texttt{not vulnerable}, and \texttt{no vulnerabilities}. Negations are handled explicitly, and mixed or conflicting evidence yields an \texttt{UNKNOWN} classification. The reported study configuration corresponds most closely to \texttt{verdict\_first} with the \texttt{full} parser. In the updated framework, \texttt{strict} and \texttt{structured} modes are exposed as explicit ablations so parser sensitivity can be measured rather than hidden. Full excerpts are shown in Appendix~\ref{appendix:label-parser}.

The main study fixes parser mode across all model--prompt comparisons. This controls parser behavior when comparing prompting strategies, but it does not make the absolute scores independent of extraction choices: the \texttt{full} parser still includes lexical fallback. We therefore treat parsing as part of the evaluation setup and expose \texttt{strict}, \texttt{structured}, and \texttt{full} modes as ablation controls. The reported prompt-strategy comparisons should be read under that fixed parsing configuration.
\begin{figure}[t]
    \centering
    \includegraphics[width=.99\columnwidth]{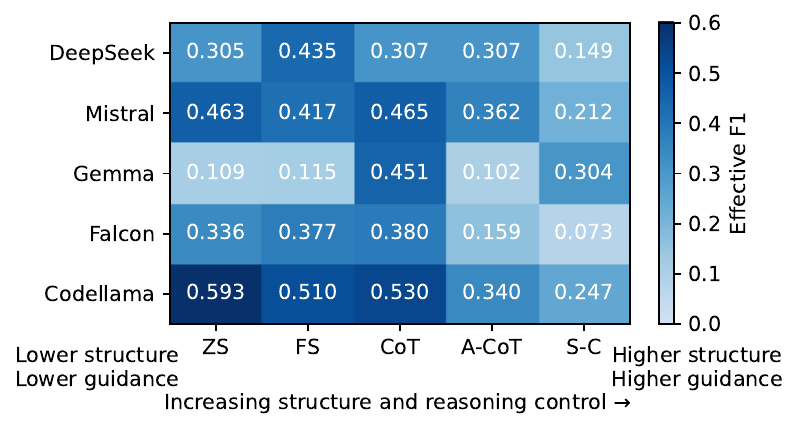}\vspace{-2mm}
    \caption{Effective F1 across model–prompt combinations. Performance improves with moderate structure (FS, CoT) but degrades under higher-structure prompts (A-CoT, S-C), indicating a non-monotonic trade-off.}
    \label{fig:sensitivity-heatmap}\vspace{-3mm}
\end{figure}

\subsection{Evaluation Metrics}
\label{sec:eval-metrics}

We evaluate each model–prompt configuration using accuracy, precision, recall, abstention rate, coverage, standard F1, and effective F1. Because vulnerability detection is an asymmetric classification task, accuracy and F1 alone are insufficient: a model may perform well on answered samples while abstaining on a substantial portion of inputs, reducing its practical utility.

Each prediction is mapped to one of six outcome categories: true positive (TP), true negative (TN), false positive (FP), false negative (FN), incorrect, and unknown false negative (UnFN). TP and TN denote correct \vul and \safe predictions, respectively. FP denotes code incorrectly classified as \vul, and FN denotes vulnerable code incorrectly classified as \safe. UnFN captures vulnerable samples for which the model fails to produce a definitive \vul verdict, including abstentions and unresolved outputs. Incorrect captures unresolved or invalid outputs on non-vulnerable samples, including protocol violations, refusals, or responses that cannot be mapped to either \safe or \vul.

Accuracy is defined as $\frac{TP + TN}{TP + TN + FP + FN + UnFN}$, precision as $\frac{TP}{TP + FP}$, and recall as $\frac{TP}{TP + FN + UnFN}$. Incorrect outputs are excluded from the accuracy denominator because they reflect protocol non-compliance rather than a committed \safe/\vul classification; their impact is instead captured through abstention, coverage, and effective F1. Recall is defined conservatively, as abstentions on vulnerable samples correspond to missed detections. Standard F1 is the harmonic mean of precision and recall. We define abstention rate as $\frac{Incorrect + UnFN}{TP + TN + FP + FN + UnFN + Incorrect}$ and coverage as one minus abstention rate. Our primary operational metric is effective F1, defined as the product of standard F1 and coverage, which penalizes configurations that achieve strong conditional performance only by abstaining on many inputs.

Effective F1 is not intended as a complete deployment cost model and scales F1 by coverage to penalize unresolved outputs, while recall separately treats abstentions on vulnerable samples as missed detections via UnFN. In practice, abstained samples may be routed to fallback analyzers or human review, whereas false negatives may pass silently. Different deployments may therefore weight false positives, false negatives, and abstentions differently. We use effective F1 as a consistent reporting measure across configurations.

\section{Evaluation and Results}
\label{sec:evaluation-results}
\subsection{Aggregate Performance}

Figure~\ref{fig:sensitivity-heatmap} summarizes effective-F1 differences across all model and template combinations; the full standard-F1 and effective-F1 table is provided in Appendix~\ref{appendix:supplementary-results} (Table~\ref{tab:f1_effective}). Three structural patterns emerge. The absolute scores should not be interpreted as evidence that these models are deployment-ready vulnerability detectors. Many configurations operate near chance precision, and even the strongest configurations remain limited by snippet-level labels and missing program context. The central result is not high detection accuracy, but the magnitude and direction of behavioral changes induced by prompt-template choice under fixed evaluation conditions.

First, standard and effective F1 diverge substantially only when abstention is elevated: Falcon's mean standard F1 (0.430) matches Mistral's, but its mean effective F1 (0.265) is 0.119 points lower, driven entirely by a mean abstention rate of 42.26\% that persists across all five strategies. Second, Mistral serves as the stability baseline: it achieves the smallest cross-strategy F1 range (0.103), indicating low sensitivity to prompt formulation. Third, Gemma serves as the prompt-sensitivity baseline: it exhibits the largest cross-strategy F1 range (0.398), with standard F1 spanning from 0.102 under A-CoT to 0.499 under self-consistency, showing prompt choice can fundamentally determine whether a model detects vulnerabilities at all.  We report deterministic point estimates under a fixed seed and decoding configuration. Because this study focuses on controlled comparisons rather than repeated stochastic sampling, we emphasize large and directionally consistent differences across prompt templates. Estimating run-to-run, template-to-template, and backend-level uncertainty is our future work.

\subsection{Effect of Prompting Strategy}

\begin{figure}[t]
    \centering
    \includegraphics[width=.99\columnwidth]{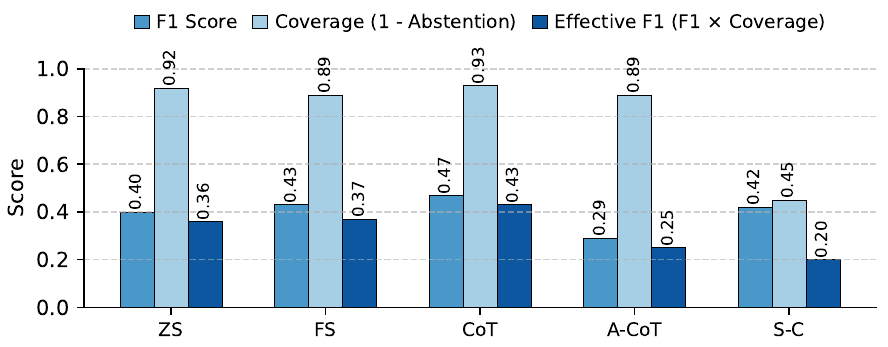}\vspace{-2mm}
    \caption{F1, coverage, and effective F1 by prompt template. High coverage is consistent across strategies, but variation in F1 drives overall effectiveness, which peaks for FS and CoT and declines for A-CoT and S-C.}
    \label{fig:decomposing-performance}\vspace{-3mm}
\end{figure}

In the following, we treat the evaluated prompt-template instance as the primary independent variable and examine its impact on recall, abstention, and effective F1 across models. Figure~\ref{fig:decomposing-performance} complements the tabular results by decomposing mean performance into standard F1, coverage, and effective F1 across the evaluated prompt-template instances. Results are drawn from Figure~\ref{fig:decomposing-performance}, Table~\ref{tab:recall}, and  Table~\ref{tab:f1_effective}; per-cell accuracy, precision, and recall appear in Appendix~\ref{appendix:metrics} (Table~\ref{tab:appendix-metrics}), and abstention rates are Table~\ref{tab:abstention-rate}. The output-protocol ablation in Section~\ref{sec:ablation} shows that these strategy-level differences are not explained solely by verdict placement.

\begin{table}[t]
\caption{Abstention rate (\%) by model and prompt template. Means and ranges (max--min) are reported across templates for each model and across models for each template.}
\label{tab:abstention-rate}\vspace{-3mm}
\centering
\scalebox{0.82}{\begin{tabular}{lccccccc}
\hline
\textbf{Model}
& {\bf ZS}
& {\bf FS}
& {\textbf{CoT}}
& {\textbf{A-CoT}}
& {\textbf{S-C}}
& {\textbf{Mean}}
& {\textbf{Range}} \\
\hline
DeepSeek   & 1.25\% & 14.32\% & 1.12\% & 0.54\% & 60.96\% & 15.64\% & 60.42\% \\
Mistral    & 0.71\% & 1.51\%  & 0.08\% & 0.25\% & 51.07\% & 10.72\% & 50.99\% \\
Gemma      & 0.00\% & 0.00\%  & 0.00\% & 0.00\% & 39.15\% & 7.83\%  & 39.15\% \\
Falcon     & 31.66\%& 27.99\% & 27.49\%& 48.19\%& 75.96\% & 42.26\% & 48.47\% \\
CodeLlama  & 5.47\% & 11.90\% & 7.70\% & 4.48\% & 49.00\% & 15.71\% & 44.52\% \\
\hline
\textbf{Mean}  & 7.82\% & 11.15\% & 7.28\% & 10.69\% & 55.23\% &        &        \\
\textbf{Range} & 31.66\%& 27.99\% & 27.49\%& 48.19\% & 36.81\% &        &        \\
\hline
\end{tabular}}\vspace{-3mm}
\end{table}

Figure~\ref{fig:decomposing-performance} reveals that variation in effective performance is driven primarily by changes in standard F1 rather than coverage. Across ZS, FS, CoT, and A-CoT, coverage remains high and relatively stable, while F1 varies substantially, causing effective F1 to track F1 closely in most cases. The main exception was under configurations that induce higher abstention, where coverage drops and effective F1 diverges from standard F1. This decomposition also shows a non-monotonic effect of prompt structure where moderate structure (few-shot and CoT) yields the highest F1 and effective F1 and more complex strategies (A-CoT and our A-CoT-based self-consistency configuration) degrade performance, either through recall reduction or increased abstention. These results indicate that increasing prompt complexity does not consistently improve detection quality, but instead shifts the balance between recall and coverage.

\BfPara{Quantitative Analysis} 
{\em The evaluated CoT template} achieves the highest mean F1 (0.465) and mean effective F1 (0.427), with a low mean abstention rate of 7.28\%. Its cross-model recall range (0.458) is substantially lower than that of zero-shot (0.805) and few-shot (0.641), indicating that it combines strong average performance with moderate cross-model stability under our setup. 

{\em The evaluated few-shot template} yields the second-highest mean F1 (0.430) and effective F1 (0.371), but exhibits high cross-model recall variability (0.641), suggesting uneven benefit across architectures. {\em The evaluated zero-shot template} further reduces structure, producing a mean F1 of 0.401 with the second-lowest abstention rate (7.82\%) but the highest cross-model recall variability (0.805), reflecting strong dependence on model-specific behavior. 

For more complex prompting, {\em our A-CoT-based self-consistency configuration} achieves a mean standard F1 (0.420) but the lowest mean effective F1 (0.197), with a gap of 0.223 points driven by a high mean abstention of 55.23\%. {\em The A-CoT template} similarly underperforms, producing the lowest mean standard F1 (0.287) despite a moderate abstention rate (10.69\%), indicating that its degradation is driven primarily by recall collapse rather than coverage loss. Recall under A-CoT falls below the corresponding CoT value for every model, with the largest drops in CodeLlama (0.685~$\rightarrow$~0.279) and Falcon (0.541~$\rightarrow$~0.223).  Implications are discussed in Section~\ref{subsec:discuss-threshold}.

\begin{rqbox}
{\bf AQ1:} 
Prompt-template choice substantially affects performance, coverage, and abstention. CoT provides the best overall balance, while A-CoT reduces recall and our A-CoT-based self-consistency configuration sharply reduces coverage through abstention. Overall, prompt effects are non-monotonic: moderate structure improves performance, but additional complexity degrades reliability via recall suppression or abstention.
\end{rqbox}

\subsection{Precision--Recall Tradeoff}

\begin{figure}[t]
    \centering
    \includegraphics[width=.99\columnwidth]{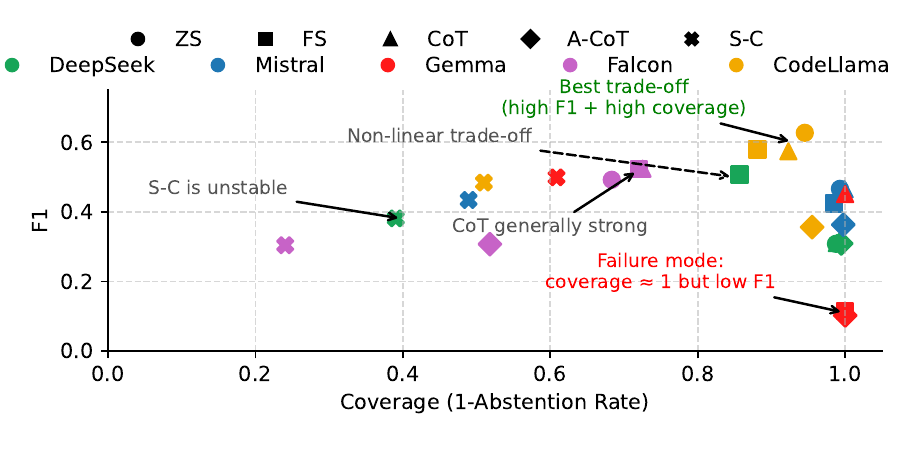}\vspace{-2mm}\vspace{-3mm}
    \caption{Coverage vs. F1 across model–prompt combinations. High coverage does not guarantee high F1; CoT provides stronger trade-offs, while S-C is unstable and some settings collapse to high-coverage, low-F1 regimes.}
    \label{fig:coverage-accuracy}\vspace{-3mm}
\end{figure}

Figure~\ref{fig:coverage-accuracy} puts all 25 model--prompt combinations in coverage--accuracy space. Selected examples illustrating the extremes of recall and abstention behavior are in Appendix~\ref{appendix:supplementary-results} (Table~\ref{tab:tradeoff}), with additional results in Appendix~\ref{appendix:graphs-tables}; full results in Appendix~\ref{appendix:metrics} (Table~\ref{tab:appendix-metrics}).

A consistent pattern across all models is that recall varies substantially while precision remains confined to a narrow range. In CodeLlama, precision holds at 0.492 across zero-shot and A-CoT while recall swings from 0.865 to 0.279, which is a difference of 0.586 points with accuracy remaining nearly flat (0.472 vs. 0.484). In Mistral, the same dynamic appears under CoT and self-consistency: precision stays within 0.003 points while recall drops from 0.423 to 0.375 and effective F1 falls from 0.465 to 0.212 due to abstention.

\begin{table}[t]
\centering
\caption{Recall performance by model and prompting strategy, with precision ranges. The bottom row shows variability (max--min) across models, highlighting sensitivity to prompt.}\label{tab:recall}\vspace{-2mm}
\scalebox{0.9}{\begin{tabular}{lccccc c}
\hline
Model & ZS & FS & CoT & A-CoT & S-C & Prec. Range \\
\hline
DeepSeek 
& 0.224 
& \cellcolor{green!20}0.497 
& 0.227 
& 0.226 
& 0.297 
& 0.487--0.531 \\

Mistral 
& \cellcolor{green!20}0.438 
& 0.359 
& 0.423 
& \cellcolor{red!15}0.280 
& 0.375 
& 0.499--0.517 \\

Gemma 
& \cellcolor{red!15}0.060 
& 0.063 
& 0.420 
& \cellcolor{red!15}0.057 
& \cellcolor{green!20}0.486 
& 0.488--0.646 \\

Falcon 
& 0.480 
& 0.540 
& \cellcolor{green!20}0.541 
& \cellcolor{red!15}0.223 
& 0.216 
& 0.492--0.517 \\

CodeLlama 
& \cellcolor{green!20}0.865 
& 0.704 
& 0.685 
& \cellcolor{red!15}0.279 
& 0.476 
& 0.491--0.495 \\
\hline
Range 
& \cellcolor{yellow!20}0.805 
& \cellcolor{yellow!15}0.641 
& 0.458 
& \cellcolor{green!10}0.223 
& \cellcolor{green!10}0.270 \\
\hline
\end{tabular}}\vspace{-3mm}
\end{table}

Gemma exhibits the strongest form of this pattern. Recall under zero-shot and A-CoT falls below 0.063, rising to 0.420 under CoT and 0.486 under self-consistency, while precision remains bounded within a 0.158-point range (Table~\ref{tab:recall}).  Across all models, precision remains tightly bounded (Table~\ref{tab:recall}), while recall spans a much wider interval. Table~\ref{tab:recall} makes this mechanism explicit: recall varies substantially across prompting strategies, whereas precision changes only marginally. This decoupling indicates that prompt-template choice primarily affects the model’s willingness to assign the \texttt{VULNERABLE} label, rather than its ability to distinguish between correct and incorrect predictions once a decision is made. In effect, prompting shifts the operating point of the detector by altering recall, not by improving precision. The implications for interpreting reported performance gains are discussed in Section~\ref{subsec:discuss-threshold}. 

This pattern is further reflected in the variability row of Table~\ref{tab:recall}. Prompting strategies such as zero-shot and few-shot exhibit large cross-model recall ranges (0.805 and 0.641, respectively), indicating that they amplify differences between model architectures. In contrast, A-CoT and S-C show reduced variability (0.223 and 0.270), but this compression arises from uniformly lower recall rather than improved consistency. Thus, reduced variability does not imply more reliable detection, but often reflects a conservative operating regime where vulnerabilities are systematically missed.

Abstention compounds this effect under our A-CoT-based self-consistency configuration. Falcon maintains precision within 0.009 points across the evaluated CoT template and this self-consistency configuration, but effective F1 collapses from 0.380 to 0.073 as abstention rises from 27.49\% to 75.96\%. This reflects a shift toward a more conservative operating regime in which both recall and coverage are suppressed, reducing operational utility despite stable precision.
Beyond individual configurations, Figure~\ref{fig:coverage-accuracy} reveals three distinct operational regimes. First, several configurations cluster in a high-coverage but low-F1 region, indicating that the model produces predictions consistently but fails to identify vulnerabilities reliably; this corresponds to {\em recall-limited behavior} where predictions are made but often incorrect. Second, the strongest configurations lie in the upper-right region, achieving both high coverage and high F1, reflecting a {\em balanced operating point} where the model both commits to predictions and maintains reasonable detection performance. Third, configurations based on our A-CoT-based self-consistency setup shift sharply leftward, indicating {\em coverage collapse due to abstention}, even when conditional precision remains stable.

This structure highlights that the trade-off between coverage and effectiveness is non-linear. Improvements in F1 do not arise from uniformly better predictions, but from shifts in the model’s operating point along the precision–recall spectrum, coupled with varying abstention behavior. In particular, movement along the horizontal axis reflects changes in abstention, while vertical variation reflects changes in recall-driven effectiveness. As a result, two configurations with similar precision can occupy very different regions of the plot and exhibit substantially different operational utility. Importantly, no configuration dominates the space, reinforcing that prompt design does not uniformly improve performance but repositions the detector within a constrained trade-off surface.

\begin{rqbox}
{\bf AQ2:}
Prompting primarily shifts the precision–recall tradeoff by altering recall rather than improving accuracy. Precision remains stable across models, while recall varies substantially, so F1 gains reflect shifts in the operating point rather than better reasoning. Accuracy and precision alone are therefore insufficient, as they miss recall- and coverage-driven effects.
\end{rqbox}

\subsection{Failure Modes}
\label{subsec:failure-modes} 

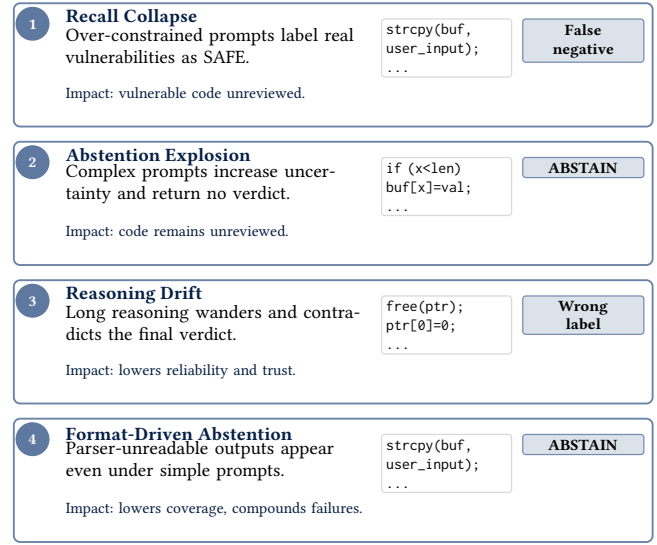
\begin{figure}[t]
\centering
\resizebox{\columnwidth}{!}{%
\begin{tikzpicture}[
    font=\sffamily,
    mode/.style={
        draw=#1!60,
        line width=0.6pt,
        rounded corners=2pt,
        minimum width=7.15cm,
        minimum height=1.38cm,
        anchor=north west
    },
    badge/.style={
        circle,
        fill=#1!65,
        text=white,
        font=\bfseries\tiny,
        minimum size=3.8mm,
        inner sep=0pt
    },
    title/.style={
        font=\bfseries\scriptsize,
        text=#1!65!black,
        anchor=west
    },
    desc/.style={
        font=\scriptsize,
        text width=3.35cm,
        align=left,
        anchor=west
    },
    code/.style={
        draw=black!15,
        fill=white,
        rounded corners=1pt,
        font=\ttfamily\tiny,
        text width=1.35cm,
        inner sep=1.5pt,
        anchor=north west
    },
    pred/.style={
        draw=#1!55,
        fill=#1!14,
        rounded corners=1pt,
        font=\bfseries\tiny,
        text width=1.25cm,
        align=center,
        inner sep=1.5pt,
        anchor=north west
    },
    impact/.style={
        font=\tiny,
        text=#1!60!black,
        text width=3.35cm,
        align=left,
        anchor=west
    }
]

\node[mode=blueE] at (0,0) {};
\node[badge=blueE] at (0.22,-0.24) {1};
\node[title=blueE] at (0.48,-0.17) {Recall Collapse};
\node[desc] at (0.48,-0.48) {Over-constrained prompts label real vulnerabilities as SAFE.};
\node[impact=blueE] at (0.48,-1.03) {Impact: vulnerable code unreviewed.};
\node[code] at (4.12,-0.18) {strcpy(buf,\\user\_input);\\...};
\node[pred=blueE] at (5.70,-0.18) {False\\negative};

\node[mode=blueE] at (0,-1.55) {};
\node[badge=blueE] at (0.22,-1.79) {2};
\node[title=blueE] at (0.48,-1.72) {Abstention Explosion};
\node[desc] at (0.48,-2.03) {Complex prompts increase uncertainty and return no verdict.};
\node[impact=blueE] at (0.48,-2.58) {Impact: code remains unreviewed.};
\node[code] at (4.12,-1.73) {if (x<len)\\buf[x]=val;\\...};
\node[pred=blueE] at (5.70,-1.73) {ABSTAIN};

\node[mode=blueE] at (0,-3.10) {};
\node[badge=blueE] at (0.22,-3.34) {3};
\node[title=blueE] at (0.48,-3.27) {Reasoning Drift};
\node[desc] at (0.48,-3.58) {Long reasoning wanders and contradicts the final verdict.};
\node[impact=blueE] at (0.48,-4.13) {Impact: lowers reliability and trust.};
\node[code] at (4.12,-3.28) {free(ptr);\\ptr[0]=0;\\...};
\node[pred=blueE] at (5.70,-3.28) {Wrong\\label};

\node[mode=blueE] at (0,-4.65) {};
\node[badge=blueE] at (0.22,-4.89) {4};
\node[title=blueE] at (0.48,-4.82) {Format-Driven Abstention};
\node[desc] at (0.48,-5.13) {Parser-unreadable outputs appear even under simple prompts.};
\node[impact=blueE] at (0.48,-5.68) {Impact: lowers coverage, compounds failures.};
\node[code] at (4.12,-4.83) {strcpy(buf,\\user\_input);\\...};
\node[pred=blueE] at (5.70,-4.83) {ABSTAIN};

\end{tikzpicture}%
}\vspace{-3mm}
\caption{Common failure modes introduced by prompt-template and output-interface choices.}
\label{fig:common-failure-modes}\vspace{-5mm}
\end{figure}

Across model--template combinations, four recurring failure modes emerge. Figure~\ref{fig:common-failure-modes} summarizes these patterns and their operational effect. Full confusion statistics appear in Appendix~\ref{appendix:metrics} (Table~\ref{tab:appendix-confusion}).

\BfPara{Mode 1: Recall Collapse under A-CoT} The A-CoT template reduces recall relative to CoT for every model, with the largest drops in CodeLlama (0.685~$\rightarrow$~0.279), Falcon (0.541~$\rightarrow$~0.223), Gemma (0.420~$\rightarrow$~0.057), and Mistral (0.423~$\rightarrow$~0.280). Accuracy changes are comparatively small in several cases, so this failure is most visible through recall and effective F1 rather than accuracy alone.

\BfPara{Mode 2: Explanation Reliability} Despite producing longer and more structured responses, reasoning-based strategies do not consistently improve recall or effective F1 relative to zero-shot (Table~\ref{tab:f1_effective}). Rationale length is therefore not evidence of improved security judgment. Systematic rationale–verdict consistency analysis requires raw model outputs and is left for future work.

\BfPara{Mode 3: Coverage Collapse under Self-Consistency} Under our A-CoT-based self-consistency, coverage drops across all models when sampled outputs disagree, falling to 51.00\% for CodeLlama, 48.93\% for Mistral, 60.85\% for Gemma, 39.04\% for DeepSeek, and 24.04\% for Falcon (Table~\ref{tab:abstention-rate}). Unlike recall collapse, abstention is explicit, but the operational effect is similar: large portions of code receive no verdict. This reflects the behavior of this majority-vote policy rather than a general property of all self-consistency variants.

\BfPara{Mode 4: Format-driven Abstention} Falcon exhibits a persistently high abstention rate (42.26\%) across all prompting strategies (Table~\ref{tab:abstention-rate}), including zero-shot (31.66\%). This indicates that failure arises not from prompt complexity but from inconsistent instruction-following that produces outputs the parser cannot resolve, compounding other failure modes by reducing usable predictions independently of strategy choice.

\begin{rqbox}
{\bf AQ3:}
Three failure modes emerge. A-CoT induces recall collapse, causing vulnerable code to be missed despite stable accuracy. Our A-CoT-based self-consistency configuration causes coverage collapse, leaving large portions of code unreviewed due to abstention. Falcon exhibits persistent format-driven abstention across prompts, reducing usable predictions. Together, these show that operational reliability depends on both recall and coverage and cannot be inferred from accuracy alone.
\end{rqbox}

\subsection{Model-Level Differences}
\begin{figure}[t]
    \centering
    \includegraphics[width=.99\columnwidth]{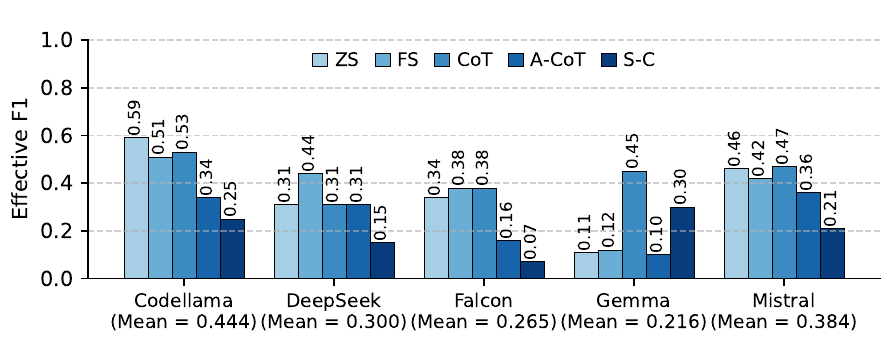}\vspace{-2mm}
    \caption{Effective F1 by model and prompt. FS and CoT typically outperform others, while A-CoT and S-C degrade performance; results remain strongly model-dependent.}
    \label{fig:performance-variability}\vspace{-5mm}
\end{figure}

This section treats model architecture as the primary variable and analyzes how different models respond to prompt variation in terms of detection capability, abstention behavior, and cross-prompt stability. Figure~\ref{fig:performance-variability} shows how effective F1 varies across models under each prompting strategy. Results are drawn from Figure~\ref{fig:performance-variability} and Table~\ref{tab:f1_effective}; per-cell abstention rates appear in Appendix~\ref{appendix:metrics} (Table~\ref{tab:abstention-rate}). 

Models separate into three behavioral groups. CodeLlama and Mistral form the high-performance tier, achieving mean F1 scores of 0.524 and 0.430, respectively, but differing in stability: CodeLlama exhibits a wider cross-strategy range (0.271), whereas Mistral shows the narrowest (0.103). Falcon occupies a distinct position, matching Mistral on mean standard F1 (0.430) but dropping to an effective F1 of 0.265 due to a persistently high abstention rate (42.26\%). DeepSeek and Gemma form a prompt-dependent tier, with performance varying substantially across strategies; Gemma in particular exhibits the largest cross-strategy F1 range (0.398).

Across all models, two distinct failure patterns emerge: recall-limited models maintain coverage but miss vulnerabilities, while coverage-limited models achieve reasonable conditional accuracy but abstain excessively. The interaction between these patterns and prompt design is discussed in Section~\ref{subsec:discuss-interaction}. 

Figure~\ref{fig:performance-variability} also reveals a consistent cross-model pattern in how prompts affect performance. Across most models, few-shot and CoT prompting yield the strongest effective F1, while A-CoT and our A-CoT-based self-consistency consistently degrade performance, either through recall reduction or increased abstention. This indicates that the effect of prompt structure is {\em directionally consistent} even though its magnitude remains model-dependent. In addition, cross-strategy variability reflects differences in how models respond to prompt formulation. Models such as Gemma and DeepSeek exhibit large performance ranges, indicating that their detection capability is highly sensitive to prompt design and can be substantially activated or suppressed depending on the template. In contrast, Mistral shows limited variation across strategies, suggesting more stable behavior under prompt changes. Falcon represents a distinct case where performance is constrained primarily by persistent abstention rather than prompt sensitivity, leading to consistently lower effective F1 despite comparable standard F1. These patterns show that prompt sensitivity is not uniform across models, but instead reflects underlying differences in how models balance recall, confidence, and abstention under varying instruction structures.

\begin{rqbox}
{\bf AQ4:}
Model–template interactions are structured and model-dependent, and no prompt-template is optimal across all models. CoT and few-shot prompting yield the strongest performance across models, while A-CoT and self-consistency consistently degrade performance through recall reduction or abstention. Models separate into distinct regimes: CodeLlama and Mistral achieve strong and stable performance, DeepSeek and Gemma are highly prompt-sensitive, and Falcon is constrained by persistent abstention. Prompt-template effects are therefore not transferable and must be validated per model family.
\end{rqbox}

\subsection{Output-Protocol Ablation}
\label{sec:ablation}

Appendix~\ref{appendix:supplementary-results} (Table~\ref{tab:ablation}) reports F1, effective F1, and abstention for all 25 model--template combinations under both output protocols on the 1,232-sample ablation subset. The protocol design is described in Section~\ref{sec:output-protocol}. In the main text, we summarize the ablation by its aggregate effect: verdict-first yields mean F1 0.403, mean effective F1 0.347, and mean abstention 9.2\%, while verdict-last yields mean F1 0.344, mean effective F1 0.248, and mean abstention 32.6\%. \emph{Overall, verdict-first outperforms verdict-last.} Switching to verdict-last consistently reduces performance, with F1 decreasing from 0.403 to 0.344, effective F1 from 0.347 to 0.248, and abstention increasing from 9.2\% to 32.6\%. Across all models except Falcon, abstention increases in every configuration, with the largest shifts in Mistral (e.g., CoT: 0.2\%~$\rightarrow$~50.0\%) and CodeLlama (e.g., A-CoT: 6.3\%~$\rightarrow$~30.0\%).

We note that \emph{there are, however, a small number of deviations.} CodeLlama Few-Shot and DeepSeek Zero-Shot show higher effective F1 under verdict-last (0.464 vs.\ 0.288 and 0.346 vs.\ 0.080, respectively). In both cases, the gain is driven by recall recovery rather than reduced abstention, indicating a shift in operating point rather than improved reasoning. On the other hand, \emph{Falcon represents a distinct regime} and shows comparatively small differences across protocols (mean effective F1: 0.281 vs.\ 0.224), consistent with its persistent format-driven abstention (Section~\ref{subsec:failure-modes}). Here, performance is constrained by instruction-following rather than protocol choice. \emph{Overall, these results support using \texttt{verdict\_first} as the default protocol} because \texttt{verdict\_last} increases unresolved outputs without improving conditional performance, reducing operational utility. While this does not fully resolve the relationship between reasoning and final decisions, it does not support the view that verdict-first CoT results are merely post-hoc rationalizations.

\section{Discussion}
\label{sec:discussion}
Prior work has shown that prompt wording and format can produce large performance swings in LLMs, and that vulnerability-detection performance varies substantially across models, datasets, and prompting setups~\cite{sclar_quantifying_2024,chatterjee_posix_2024,lin_large_2025,jiang_investigating_2025,huaFlawArtifactRethinking}. Our results sharpen this picture in a more controlled and operational setting. Rather than treating prompt sensitivity as a secondary observation within broader benchmarks, we examine concrete prompt-template instances under a fixed dataset, decoding configuration, and evaluation pipeline. This setup allows us to isolate shifts in recall, abstention, and effective coverage from changes in conditional performance, and to show that prompt choice alters operational behavior even when the underlying model and task remain unchanged. More broadly, this framing aligns with recent work arguing that robustness and reliability should be assessed using standardized, multi-metric evaluations rather than a single aggregate score~\cite{liang_holistic_2023}.

\subsection{Prompting as a Hidden Decision Threshold}
\label{subsec:discuss-threshold}

The results show that the evaluated prompt-template choices systematically alter model behavior in a way that resembles adjusting a decision threshold rather than improving underlying vulnerability understanding, consistent with prior work showing that semantically minor prompt changes can materially shift model outputs and measured performance~\cite{sclar_quantifying_2024,chatterjee_posix_2024,huaFlawArtifactRethinking}. Across all models, precision remains relatively stable while recall varies substantially, indicating that prompting shifts the model's effective decision threshold for assigning the \vul label.

This distinction is critical for interpreting performance gains. Prompt design can change the operating point of the detector without necessarily improving the quality of the underlying security reasoning~\cite{sclar_quantifying_2024,chatterjee_posix_2024,liang_holistic_2023}. Improvements in F1 therefore often reflect a more aggressive labeling policy rather than better reasoning about vulnerabilities. As a result, two prompt-template instances with similar accuracy may operate at very different points on the precision--recall curve, leading to substantially different security outcomes under the same evaluation policy.

The behavior of the A-CoT template is particularly instructive: although designed to encourage selective reasoning, it reduces recall relative to the CoT template for every model, with the largest drops in CodeLlama (0.685~$\rightarrow$~0.279), Falcon (0.541~$\rightarrow$~0.223), Gemma (0.420~$\rightarrow$~0.057), and Mistral (0.423~$\rightarrow$~0.280). This suggests that the instruction to reason selectively is interpreted as a signal to default to \safe unless strong evidence is present. One possible explanation is that A-CoT introduces an additional latent decision—whether reasoning is needed at all—which, when resolved conservatively, may bypass deeper analysis. Gemma exhibits a strong form of this pattern: under zero-shot and A-CoT, recall falls below 0.063, rising to 0.420 under CoT, reflecting a shift in operating point rather than improved semantic understanding. The output-protocol ablation in Section~\ref{sec:ablation} supports this interpretation, indicating that these effects are not explained solely by verdict placement: strategy-level differences in recall and abstention persist across both protocols.

The evaluated CoT template achieves the highest mean F1 (0.465) and mean effective F1 (0.427) across models while maintaining a low abstention rate (7.28\%), suggesting that this structured-reasoning template stabilizes outputs without substantially increasing abstention. This aligns with prior work using CoT and related methods to structure intermediate reasoning rather than change the task itself~\cite{wei2022chain,wang_self-consistency_2023,chen2025unleashing}. At the same time, the results do not establish that longer reasoning traces cause better vulnerability understanding: reasoning-oriented prompts sometimes improve final metrics and sometimes do not. Structured reasoning should therefore be treated as a controllable inductive bias rather than a reliability guarantee, particularly in security settings where prior work reports brittle reasoning and missed vulnerabilities despite fluent explanations~\cite{ullah2024llms,basic_vulnerabilities_2024,lin_large_2025}. Systematic rationale–verdict consistency analysis requires raw-output inspection and remains future work.

The failure of A-CoT is instructive rather than anomalous. More broadly, adaptive prompting has been proposed as a way to vary reasoning effort with task complexity, but the literature does not establish that such adaptation is reliable in code-security settings, especially when the model must decide whether deeper analysis is warranted~\cite{wan2023betterzeroshotreasoningselfadaptive,wan2023universalselfadaptiveprompting,r2024thinksizeadaptiveprompting,nong2024chain}. The evaluated A-CoT template was designed to mimic human review behavior by encouraging deeper reasoning on complex code, yet it underperformed the evaluated standard CoT template on both mean F1 (0.287 vs. 0.465) and mean effective F1 (0.254 vs. 0.427). Brittleness here arises not from prompting itself, but from delegating the depth-of-analysis decision to the model without explicit supervision. More robust adaptation would require explicit, verifiable triggers based on syntactic or semantic code properties, rather than relying on the model to regulate reasoning depth through natural language instruction alone.

\subsection{Abstention as a Security Risk}

Building on the threshold-shift behavior discussed above, abstention introduces a distinct and often underappreciated failure mode in vulnerability detection. More generally, evaluation work on language models has argued that system utility cannot be judged by accuracy alone when robustness, reliability, and actionability matter~\cite{liang_holistic_2023}. In code-security workflows, an abstained sample corresponds to code that receives no actionable verdict unless a downstream fallback is explicitly defined~\cite{chess2007secure,johnson2013don}.

Across models, abstention rates are substantial and highly dependent on prompt-template choice. In particular, our A-CoT-based self-consistency configuration produces widespread abstention due to disagreement across sampled outputs, reducing effective coverage even when conditional accuracy remains reasonable. This creates a silent failure mode in which the system appears accurate on evaluated samples while leaving large portions of input unprocessed. These findings indicate that abstention is not a secondary metric but a core part of system behavior. As discussed above, accuracy alone does not capture operational reliability: evaluations that omit coverage or effective F1 risk overstating model utility and obscuring risk. Falcon illustrates this concretely: despite maintaining precision within 0.009 points across the evaluated CoT template and our A-CoT-based self-consistency configuration, its effective F1 collapses from 0.380 to 0.073 as abstention rises from 27.49\% to 75.96\%, reflecting a conservative decision process that trades coverage for confidence. Our A-CoT-based self-consistency configuration amplifies this across all models, with coverage dropping to 24.04\% for Falcon and 39.04\% for DeepSeek, meaning that in the worst case fewer than one in four code samples receives any verdict.

In our implementation, self-consistency converts inter-sample disagreement into abstention when sampled outputs fail to produce a clear majority label. When models are uncertain, outputs frequently split across \safe and \vul without converging, causing majority voting to return no verdict rather than a potentially unreliable one. These observations support treating vulnerability detection as an operational reliability problem rather than a closed-world classification task, where a complete assessment requires explicit policies defining when abstention is acceptable and how abstained cases should be handled, such as escalation to static analysis tools or human review~\cite{liang_holistic_2023,chess2007secure,johnson2013don,li_iris_2025,yang_knighter_2025}.

\subsection{Model and Strategy Interaction}
\label{subsec:discuss-interaction}

Building on the abstention and threshold effects discussed above, prompt-template choice and model architecture interact in a non-uniform manner. No single evaluated template is optimal across all models, and the effectiveness of a given template depends strongly on the model's underlying capabilities, consistent with prior work showing that prompt sensitivity is structured, model-dependent, and not eliminated by scale or instruction tuning~\cite{sclar_quantifying_2024,chatterjee_posix_2024,lin_large_2025,jiang_investigating_2025}.

CodeLlama achieves the strongest mean F1 (0.524) even under zero-shot prompting, consistent with the idea that code-specialized pretraining provides stronger priors for program understanding and vulnerability-pattern recognition than general-purpose instruction tuning alone~\cite{roziere2023code,fan2023large,hou2024large}. Mistral exhibits the most consistent behavior across strategies (F1 range: 0.103), making it less sensitive to prompt formulation and more suitable for deployment scenarios requiring predictable coverage. In contrast, DeepSeek and Gemma show strong prompt dependence: DeepSeek's F1 increases from 0.308 under zero-shot to 0.507 under few-shot, while Gemma's F1 spans 0.398 across strategies, indicating that its capability is present but inconsistently activated depending on task framing.

These differences indicate that prompt sensitivity is systematic under our evaluation setup rather than random evaluation noise. Reporting a single model–prompt result therefore provides an incomplete picture: relative rankings shift with strategy choice, and a configuration that appears optimal under one prompt may underperform under another. Aggregate comparisons that fix a single strategy obscure model-specific behavior and risk overinterpreting prompt-specific results as general capability differences, especially when prompt format, scoring, and extraction choices can also influence apparent rankings~\cite{liang_holistic_2023,huaFlawArtifactRethinking,sclar_quantifying_2024}.

\subsection{Reproducibility Implications}

Building on the model-dependent prompt effects discussed above, the sensitivity of model performance to prompting strategy has direct implications for reproducibility. When the same model evaluated on the same dataset produces substantially different results under different prompts, a single reported metric does not fully characterize its behavior~\cite{sclar_quantifying_2024,chatterjee_posix_2024,huaFlawArtifactRethinking}.

This variability complicates comparisons across studies and makes it difficult to disentangle methodological gains from prompt-induced effects unless evaluation conditions are explicitly standardized and reported~\cite{liang_holistic_2023,sclar_quantifying_2024,chatterjee_posix_2024}. Reported improvements may reflect differences in prompting, scoring, or output handling rather than genuine advances in vulnerability detection capability~\cite{huaFlawArtifactRethinking}. As shown earlier, prompt-template choice can materially alter recall and coverage, so studies that report a single configuration risk overinterpreting prompt-specific behavior as model-level capability.

Reliable evaluation therefore requires treating prompting as a controlled variable. The cross-strategy F1 ranges observed in Table~\ref{tab:f1_effective}, spanning 0.398 for Gemma and 0.271 for CodeLlama, demonstrate that a single reported metric does not characterize model behavior. Two studies evaluating the same model on the same dataset with different prompts can therefore reach opposite conclusions. Multi-strategy evaluation provides a more complete characterization and reduces the risk of overinterpreting single-configuration results. The output-protocol ablation in Section~\ref{sec:ablation} further supports that the strategy-level effects are not explained solely by verdict placement or parser heuristics, supporting the validity of cross-strategy comparisons under the \PA\ framework.

Although \PA\ holds datasets, decoding parameters, and parsing rules constant to isolate prompt effects, the results show that evaluation design choices themselves influence measured performance, consistent with broader work emphasizing that benchmark conclusions depend materially on how scenarios, metrics, and scoring procedures are defined~\cite{liang_holistic_2023,huaFlawArtifactRethinking}. Strict output constraints reduce ambiguity but increase abstention when models deviate from the expected format, as evidenced by Falcon's mean abstention of 42.26\% persisting across all five strategies regardless of prompt complexity (Appendix~\ref{appendix:metrics}, Table~\ref{tab:abstention-rate}). Aggregation under self-consistency improves conditional F1 (0.420) while collapsing mean effective F1 to 0.197, showing that aggregation strategies can improve standard metrics while reducing operational coverage. These dependencies indicate that the evaluation harness is part of the causal pathway by which prompting affects outcomes. Making these design choices explicit, as \PA\ does through its fixed pipeline and ablation controls, is essential for interpreting results and enabling meaningful cross-study comparison.

\subsection{Evaluation Implications}
\label{subsec:discuss-deployment}

Building on the reproducibility and interaction effects discussed above, the 25 model-template combinations separate into three reliability tiers under this evaluation setup. \textbf{High-performing combinations:} CodeLlama and Mistral paired with the evaluated zero-shot or CoT templates provide the strongest and most stable operational performance. \textbf{Moderately reliable combinations:} Falcon under the evaluated few-shot or CoT templates, and DeepSeek under the evaluated few-shot template, are usable but sensitive to coverage changes. \textbf{Unstable combinations:} Any model under our A-CoT-based self-consistency configuration, DeepSeek under the evaluated zero-shot or adaptive reasoning templates, and Gemma under the evaluated zero-shot or A-CoT templates, all of which exhibit large performance swings or severe abstention.

These tiers reinforce that prompt-template selection must be validated per model family rather than uniformly. Moreover, the findings translate into several practical considerations for LLM-based vulnerability detection systems. First, evaluation should report coverage and abstention with F1, as standard metrics do not capture operational reliability; effective F1 and abstention rates are necessary to assess real-world performance. Second, certain prompt-template designs should be used with caution; the evaluated adaptive CoT template and our A-CoT-based self-consistency introduce failure modes such as recall suppression and coverage collapse, making similar designs unsuitable for smaller models without careful validation; this aligns with broader findings that reasoning-oriented prompts can behave unpredictably in code-security tasks despite their success on generic reasoning benchmarks~\cite{wang_self-consistency_2023,ullah2024llms,basic_vulnerabilities_2024}.

Prompt-template selection should therefore be model-specific rather than uniform, as prior work and our results show that prompt sensitivity is structured and architecture-dependent rather than a generic nuisance variable~\cite{sclar_quantifying_2024,chatterjee_posix_2024,lin_large_2025}. Code-oriented models often perform well under simpler prompts, while less specialized models benefit from structured reasoning. Given this sensitivity, each model–prompt combination should be evaluated on representative data prior to deployment through a systematic prompt audit. In addition, deployment pipelines must define explicit handling policies for abstention. Samples that receive no verdict should be routed to fallback systems or human review, as treating them as safe introduces systematic risk equivalent to missed detections~\cite{chess2007secure,johnson2013don,li_iris_2025,yang_knighter_2025}.

Finally, evaluation and optimization should prioritize recall and effective F1 over precision. As shown earlier, precision remains relatively stable across configurations, while recall and coverage determine security outcomes.

\subsection{Limitations and Future Work}
\label{sec:limitations}

This study examines prompt sensitivity using open-weight language models with fewer than 7B parameters, reflecting practical hardware constraints and an emphasis on fully reproducible, local experiments. While this enables controlled comparisons across strategies, it limits generalization to larger or proprietary models, which may exhibit different instruction-following behavior due to scale or extensive post-training. The goal is not to establish absolute performance ceilings, but to analyze how prompt formulation influences model behavior when other factors are held constant. Within this regime, the observed differences reflect prompt-induced effects, even if their magnitude may vary at larger scales.

The task is framed as binary \safe/\vul classification over code snippets from real CVE records. This abstraction simplifies security judgments that, in practice, depend on broader program context, exploitability conditions, and severity considerations. For instance, some snippets lack sufficient information for a definitive decision, and model abstentions or inconsistencies may reflect legitimate semantic uncertainty than failure. Thus, binary classification should be viewed as a controlled proxy for studying prompt sensitivity rather than a complete model of vulnerability analysis.

Each prompting strategy is instantiated with a single prompt template, so the results should be interpreted as strategy–template measurements rather than universal claims about all possible zero-shot, few-shot, CoT, A-CoT, or self-consistency prompts. Alternative templates may shift absolute scores and, in some cases, cross-strategy rankings, although the output-protocol ablation in Section~\ref{sec:ablation} suggests that the main trends are not driven solely by verdict placement. CVEfixes commit-level labels also carry residual noise from multi-stage patches and context-dependent vulnerabilities; these constraints apply uniformly across all conditions and are unlikely to fully explain the differential prompt effects observed. Finally, all experiments use temperature~0.2, reflecting near-deterministic behavior; systems deployed at higher temperatures may exhibit additional variance not captured here.

Taken together, these considerations highlight broader evaluation risks that extend beyond prompt design. The CVEfixes dataset is publicly available and may have been included in the training data of some models. Although this cannot be verified, it introduces a risk of data leakage and potential bias in the evaluation, a concern widely acknowledged in benchmark-based studies where overlap between training and test data is difficult to fully rule out.

\BfPara{Future Work}
Future work will strengthen \ours along three axes: realism, task coverage, and robustness. We plan to validate our findings on additional real-world datasets (e.g., \textbf{Big-Vul}~\cite{fan_cc_2020}) to assess whether prompt-driven behaviors persist across diverse projects, languages, and labels. We will extend beyond coarse classification to support \emph{vulnerability-type/CWE}, \emph{localization} to vulnerable functions or lines, and \emph{fix-oriented suggestions} in structured outputs aligned with security review workflows~\cite{li_vuldeepecker_2018,fu_linevul_2022,pearce2023examining}. We also plan to evaluate whether model explanations can be converted into normalized vulnerability-report artifacts~\cite{althebeiti2025mujaz,
althebeiti2025enhancing}.

To improve reliability, we will incorporate lightweight static-analysis cues (e.g., unsafe API flags and simple taint or dataflow indicators) and introduce a controlled prompt-perturbation suite to quantify stability across templates, formatting variations, and added context, reporting variance rather than single-prompt outcomes~\cite{sclar_quantifying_2024,chatterjee_posix_2024}. We will also evaluate deployment relevance through \emph{human-in-the-loop} studies (e.g., triage speed and error reduction), complemented by scalable \emph{LLM-as-a-judge} rubrics for explanation, localization, and fix quality, and benchmark prompting against stronger alternatives, including \emph{fine-tuned} models and \emph{tool-augmented} pipelines that combine static analyzers with LLM reasoning~\cite{du_generalization-enhanced_2024,li_iris_2025,yang_knighter_2025,johnson2013don}.

\section{Conclusion}
\label{sec:conclusion}
We studied how prompt formulation affects LLM-based vulnerability classification and introduced \ours, a framework that fixes the dataset, decoding configuration, output protocol, parser, and metrics while varying prompt structure. Across five prompt templates and multiple open-weight models on a CVE-derived binary \safe/\vul task, prompt choice changed recall, abstention, and effective F1 enough to change model rankings and evaluation conclusions.

The main lesson is that prompt-template choice behaves like part of the decision policy. It can shift a model toward higher recall, higher abstention, or more conservative labeling without changing the underlying model. Single-prompt scores therefore give an incomplete account of reliability. \ours provides a controlled way to measure this variance and to compare models under the same data, parsing, and reporting assumptions. We conclude that prompt sensitivity should be reported as part of LLM-based security evaluation. Coverage, abstention, and prompt-induced variability are necessary context for judging whether a classifier can provide consistent, actionable outputs in security-critical workflows.


\appendix

\section*{Ethical Considerations}
This research evaluates prompt sensitivity in LLM-based vulnerability detection using publicly disclosed CVE records. Stakeholders include the research team, security practitioners, model developers, open-source maintainers whose code appears in CVE records, and the broader security community.

The primary ethical benefit is exposing reliability limitations in LLM-based detection tools before deployment, enabling defenders to make informed decisions about operational constraints and failure modes. While adversaries could exploit knowledge of model weaknesses to select prompts that suppress detection, this information asymmetry already exists---adversaries can conduct similar experiments privately. Publication enables transparent evaluation and responsible deployment practices, with benefits to defenders substantially outweighing potential adversarial advantages.

No human subjects participated in this research and all analyzed code derives from publicly disclosed CVE records already part of security disclosure processes. We also perform no novel vulnerability discovery or responsible disclosure. The framework and findings are publicly available, focusing on open-weight models to ensure reproducibility without proprietary access or substantial computational resources. We conclude that publication serves the public interest by documenting systematic weaknesses that could lead to missed vulnerabilities if these tools are deployed without understanding their limitations. Transparent documentation of coverage-accuracy tradeoffs and failure modes enables more responsible development and deployment of LLM-based security tools.

\section*{Appendix}

\section{Prompt Templates and Output Protocol}
\label{appendix:prompts-parser}

We provide abbreviated versions of the prompt templates used in \ours. These excerpts document the exact interface between the experiment runner and the models.

\subsection{Output Protocol Instructions}

All prompting strategies share the same task-specific verdict instructions appended by the experiment runner after the strategy prompt. The selected output protocol controls only where the final verdict must appear; the prompting strategy controls whether and how reasoning is encouraged. Representative protocol suffixes are shown below.

\textbf{Verdict-First Protocol}
\begin{tcolorbox}[
  enhanced,
  breakable,
  colback=gray!5,
  colframe=gray!60,
  boxrule=0.99pt,
  arc=2pt,
  left=3pt,
  right=3pt,
  top=3pt,
  bottom=3pt
]
\ttfamily\footnotesize
TASK: Classify the code's security.\\
On the FIRST LINE ONLY, output exactly one of these words:
SAFE or VULNERABLE.\\
Do not add any other words, punctuation, or symbols on that first line.\\
If you include any explanation, it must begin on the SECOND line.
\end{tcolorbox}

\textbf{Verdict-Last Protocol}
\begin{tcolorbox}[
  enhanced,
  breakable,
  colback=gray!5,
  colframe=gray!60,
  boxrule=0.99pt,
  arc=2pt,
  left=3pt,
  right=3pt,
  top=3pt,
  bottom=3pt
]
\ttfamily\footnotesize
TASK: Classify the code's security.\\
If you include any explanation, it must appear before the final verdict.\\
On the FINAL LINE ONLY, output exactly one of these words:
SAFE or VULNERABLE.\\
Do not add any other words, punctuation, or symbols on that final line.
\end{tcolorbox}

The label parser described in Section~\ref{sec:label-parser} treats the protocol-defined verdict line as the primary target: the first non-empty line for \texttt{verdict\_first} and the last non-empty line for \texttt{verdict\_last}.

\subsection{Zero-Shot Prompt (Excerpt)}

The zero-shot template adds a minimal instruction on how to approach the
classification, without examples or explicit reasoning steps. The runner appends the selected output-protocol suffix after this strategy text:

\begin{tcolorbox}[
  enhanced,
  breakable,
  colback=gray!5,
  colframe=gray!60,
  boxrule=0.99pt,
  arc=2pt,
  left=3pt,
  right=3pt,
  top=3pt,
  bottom=3pt
]
\ttfamily\footnotesize
You are a secure code auditor. Examine the following code and determine whether it is secure or vulnerable to potential exploits.\\
Do not use step-by-step reasoning unless the selected prompt strategy explicitly asks for it.\\
If you include any explanation, keep it brief.\\
Briefly consider issues such as buffer overflows, injections, improper validation, and other common weaknesses.\\

<CODE\_SNIPPET\_HERE>
\end{tcolorbox}

\subsection{Few-Shot Prompt (Excerpt)}

The few-shot strategy extends the zero-shot  with compact labeled
examples. In the current implementation, these are synthetic CVE-style before/after function snippets designed to resemble function-level patch data without copying exact dataset rows:

\begin{tcolorbox}[
  enhanced,
  breakable,
  colback=gray!5,
  colframe=gray!60,
  boxrule=0.99pt,
  arc=2pt,
  left=3pt,
  right=3pt,
  top=3pt,
  bottom=3pt
]
\ttfamily\footnotesize
Use the examples below as prior knowledge for classifying code security.\\
First, review the examples and their labels. Then analyze the new code snippet.\\

Do not use step-by-step reasoning unless the selected prompt strategy explicitly asks for it.\\
If you include any explanation, keep it brief.\\

Examples:\\
Example 1 (before patch):\\
Code:\\
int parse\_message(const unsigned char *buf, size\_t len) \{\\
\hspace*{1em}unsigned int msg\_len;\\
\hspace*{1em}char header[256];\\
\\
\hspace*{1em}if (len < 2) \{\\
\hspace*{2em}return -1;\\
\hspace*{1em}\}\\
\\
\hspace*{1em}msg\_len = ((unsigned int)buf[0] << 8) | buf[1];\\
\hspace*{1em}memcpy(header, buf + 2, msg\_len);\\
\hspace*{1em}return 0;\\
\}\\
Label: VULNERABLE\\

Example 2 (after patch):\\
Code:\\
int parse\_message(const unsigned char *buf, size\_t len) \{\\
\hspace*{1em}unsigned int msg\_len;\\
\hspace*{1em}char header[256];\\
\\
\hspace*{1em}if (len < 2) \{\\
\hspace*{2em}return -1;\\
\hspace*{1em}\}\\
\\
\hspace*{1em}msg\_len = ((unsigned int)buf[0] << 8) | buf[1];\\
\hspace*{1em}if (msg\_len > len - 2 || msg\_len > sizeof(header)) \{\\
\hspace*{2em}return -1;\\
\hspace*{1em}\}\\
\\
\hspace*{1em}memcpy(header, buf + 2, msg\_len);\\
\hspace*{1em}return 0;\\
\}\\
Label: SAFE\\

Now analyze this code from a security perspective:\\
<CODE\_SNIPPET\_HERE>
\end{tcolorbox}

\subsection{Chain-of-Thought Prompt (Excerpt)}

The CoT strategy encourages structured reasoning and uses a protocol-aware placement hint. The runner inserts one of the following hints depending on the selected output protocol: (i) ``Give the required verdict first, then explain your reasoning starting on the second line'' for \texttt{verdict\_first}; or (ii) ``Reason step by step first, then place the final verdict on the last line'' for \texttt{verdict\_last}.

\begin{tcolorbox}[
  enhanced,
  breakable,
  colback=gray!5,
  colframe=gray!60,
  boxrule=0.99pt,
  arc=2pt,
  left=3pt,
  right=3pt,
  top=3pt,
  bottom=3pt
]
\ttfamily\footnotesize
You are a secure code auditor. Analyze the following code step by step, carefully reasoning about potential security vulnerabilities such as buffer overflows, injections, improper validation, race conditions, and other common issues.\\

<PLACEMENT\_HINT>\\

Consider at least these factors in your reasoning:
\begin{enumerate}
    \item Inputs and trust boundaries
    \item Validation and sanitization
    \item Memory safety and resource management
    \item Injection, race, and logic risks
\end{enumerate}

Code:\\
<CODE\_SNIPPET\_HERE>
\end{tcolorbox}

\subsection{Adaptive Chain-of-Thought (Excerpt)}

The A-CoT strategy instructs the model to decide when
stepwise reasoning is necessary and, like CoT, uses a protocol-aware placement hint:

\begin{tcolorbox}[
  enhanced,
  breakable,
  colback=gray!5,
  colframe=gray!60,
  boxrule=0.99pt,
  arc=2pt,
  left=3pt,
  right=3pt,
  top=3pt,
  bottom=3pt
]
\ttfamily\footnotesize
You are a secure code auditor. Determine whether the following code is SAFE or VULNERABLE.\\

Adjust the depth of your reasoning to the code:
\begin{enumerate}
    \item If the code is straightforward and obviously SAFE or VULNERABLE, keep any explanation brief.
    \item If the code uses pointer arithmetic, raw memory operations, manual resource management, or complex input handling, reason through the risks step by step.
    \item <PLACEMENT\_HINT>
\end{enumerate}

Code:\\
<CODE\_SNIPPET\_HERE>
\end{tcolorbox}

\subsection{Self-Consistency Strategy}

The self-consistency strategy uses the same adaptive CoT template as A-CoT, but relies on the experiment runner to issue multiple completions under the active output protocol. Five independent responses are collected per snippet by default. Each vote is parsed individually using the selected parser mode and then aggregated by majority vote. If no label receives a true majority over the requested samples, the final result is recorded as \texttt{UNKNOWN}.

\section{Label Parsing Logic}
\label{appendix:label-parser}

This section summarizes the parser modes used to convert raw
model output into \safe, \vul, or \texttt{UNKNOWN}.
The full implementation appears in \texttt{evaluation/label\_parser.py}.

\subsection{Parser Modes}

The parser supports three modes:
\begin{itemize}[leftmargin=*,itemsep=-0.1mm]
    \item \textbf{Strict}: protocol-position check only.
    \item \textbf{Structured}: strict parsing plus explicit verdict-line patterns.
    \item \textbf{Full}: structured parsing plus whole-response lexical fallback.
\end{itemize}
All three modes begin by normalizing away empty lines and then applying the protocol-specific strict check.

\subsection{Tier 1: Protocol-Position Strict Check}

\begin{lstlisting}
def _strict_line_label(lines, output_protocol):
    if not lines:
        return "unknown", None

    idx = 0 if output_protocol == "verdict_first" else -1
    target_line = lines[idx]
    parts = target_line.split()
    if len(parts) == 1:
        label = _normalize_token_label(parts[0])
        if label:
            tier = (
                "strict_first_line"
                if output_protocol == "verdict_first"
                else "strict_last_line"
            )
            return label, tier

    return "unknown", None
\end{lstlisting}

This tier accepts only the protocol-defined verdict location. Minor punctuation is normalized away before testing the token, so values such as \safe still map to \safe.

\subsection{Tier 2: Explicit Verdict Markers}

\begin{lstlisting}[basicstyle=\ttfamily\footnotesize]
_EXPLICIT_VERDICT_PATTERNS = [
    re.compile(
        r"^(final answer|answer|classification|"
        r"verdict|label|conclusion)"
        r"\s*[:\-]?\s*(?:is\s+)?"
        r"(safe|vulnerable)\s*[.!]?\s*$",
        re.IGNORECASE,
    ),
    re.compile(
        r"^(?:the\s+)?(?:final\s+answer|answer|"
        r"classification|verdict)\s+is\s+"
        r"(safe|vulnerable)\s*[.!]?\s*$",
        re.IGNORECASE,
    ),
    re.compile(
        r"^(therefore|thus|overall|ultimately|"
        r"in conclusion),?\s+"
        r"(?:the\s+code\s+is\s+)?"
        r"(safe|vulnerable)\s*[.!]?\s*$",
        re.IGNORECASE,
    ),
    re.compile(
        r"^(?:the\s+code|this\s+code)\s+is\s+"
        r"(safe|vulnerable)\s*[.!]?\s*$",
        re.IGNORECASE,
    ),
]
\end{lstlisting}

\begin{lstlisting}[basicstyle=\ttfamily\scriptsize]
for ln in reversed(lines):
    for pattern in _EXPLICIT_VERDICT_PATTERNS:
        m = pattern.search(ln)
        if m:
            explicit_label = m.groups()[-1].lower()
            return explicit_label
\end{lstlisting}

Scanning bottom up prioritizes stable final verdicts when the model writes a longer explanation before committing to a label.

\subsection{Tier 3: Contextual Keyword Scan}

\begin{lstlisting}
lowered = text.lower()

has_not_safe      = re.search(r"\bnot\s+safe\b", lowered)
has_unsafe        = re.search(r"\bunsafe\b", lowered)
has_vulnerable    = re.search(r"\bvulnerable\b", lowered)
has_vulnerability = re.search(r"\bvulnerabilit(?:y|ies)\b", lowered)
has_exploitable   = re.search(r"\bexploitable\b", lowered)
has_at_risk       = re.search(r"\bat\s+risk\b", lowered)

has_safe_word      = re.search(r"\bsafe\b", lowered)
has_secure_word    = re.search(r"\bsecure\b", lowered)
has_not_vulnerable = re.search(r"\bnot\s+vulnerable\b", lowered)
has_no_vuln        = re.search(
    r"\bno\s+(known\s+)?vulnerabilit(?:y|ies)\b", lowered
)
\end{lstlisting}

\begin{lstlisting}
if vulnerable_signal and not positive_safe:
    return "vulnerable", "lexical_fallback"
if positive_safe and not vulnerable_signal:
    return "safe", "lexical_fallback"
return "unknown", None
\end{lstlisting}

This tier is available only in \texttt{full} mode. It helps recover verdicts from freer outputs, but conflicting or mixed evidence is intentionally mapped to \texttt{UNKNOWN}.

\section{Generation Settings and Configurations}
\label{appendix:gen-settings}

These settings match the YAML configuration used across all experiments.  
Each parameter controls a specific aspect of the model's decoding behavior.

\begin{itemize}[leftmargin=*]
    \item \textbf{Temperature: 0.2}  
    The temperature controls how much randomness is introduced when selecting
    the next token. Lower values make the model more deterministic and reduce
    the chance of drifting away from the expected output format. A value of
    0.2 helps keep the classification stable under the selected output protocol while still allowing
    the model to produce natural reasoning when the prompt strategy requests it.

    \item \textbf{Top-p: 0.9}  
    Top-p sampling (also called nucleus sampling) restricts token choices to
    the smallest possible set whose cumulative probability mass reaches the
    specified threshold. Using \(p = 0.9\) limits low-probability transitions
    without forcing fully greedy decoding, which helps maintain consistency
    over models.

    \item \textbf{Top-k: 40}  
    Top-k further constrains the sampling distribution by limiting each step
    to the 40 most likely next tokens. This avoids rare or unstable tokens
    that could violate the selected output protocol or introduce noisy reasoning.

    \item \textbf{Max new tokens: 250}  
    This caps the number of tokens a model may generate in
    response to a prompt. Although the classification itself may appear on either the first or final line depending on the selected output protocol, CoT and adaptive prompts may produce additional reasoning.

    \item \textbf{Repetition penalty: 1.0}  
    This penalty discourages the model from repeating the same
    phrase or token sequence. Using 1.0 keeps behavior
    consistent across models and avoids introducing artificial bias into the
    reasoning style.

    \item \textbf{Frequency and presence penalties: 0.0}  
    These penalties adjust how the model weighs tokens it has already used.
    Setting both to zero prevents manipulating the output
    structure in unintended ways. This helps preserve reproducibility when
    applying the selected output protocol.

    \item \textbf{Seed: 42}  
    The seed ensures that sampling decisions are repeatable. Fixing the seed
    allows multiple runs to produce comparable output, which is important for evaluating the self-consistency and ensuring differences between prompts are genuine and not due to random sampling.

    \item \textbf{Self-consistency samples: 5}  
    For the self-consistency, the model generates five independent
    completions for each snippet. \ours parses each completion
    separately and applies majority voting. Using five samples provides a
    reasonable balance between stability and computational cost.
\end{itemize}

\section{Evaluation Metrics}
\label{appendix:metrics}

Section~\ref{sec:eval-metrics} summarizes the metrics in the main text; this appendix gives the complete formulation used for reporting results. Because prompt-induced abstention plays a central role in LLM-based vulnerability detection, standard classification metrics alone are not sufficient. The definitions below are designed to capture both predictive correctness and operational coverage.

\begin{table*}[htbp]
\centering
\caption{Standard F1 and effective F1 by model and prompt template.}
\label{tab:f1_effective}\vspace{-3mm}
\small
\setlength{\tabcolsep}{5pt}
\begin{tabular}{lcccccccccccccc}
\toprule
Model & \multicolumn{2}{c}{ZS} & \multicolumn{2}{c}{FS} & \multicolumn{2}{c}{CoT} & \multicolumn{2}{c}{A-CoT} & \multicolumn{2}{c}{S-C} & \multicolumn{2}{c}{Mean} & \multicolumn{2}{c}{Range} \\
\cmidrule(lr){2-3} \cmidrule(lr){4-5} \cmidrule(lr){6-7} \cmidrule(lr){8-9} \cmidrule(lr){10-11} \cmidrule(lr){12-13} \cmidrule(lr){14-15}
 & F1 & Eff. & F1 & Eff. & F1 & Eff. & F1 & Eff. & F1 & Eff. & F1 & Eff. & F1 & Eff. \\
\midrule
DeepSeek  & 0.308 & 0.305 & 0.507 & 0.435 & 0.310 & 0.307 & 0.309 & 0.307 & 0.381 & 0.149 & 0.363 & 0.301 & 0.199 & 0.286 \\
Mistral   & 0.466 & 0.463 & 0.424 & 0.417 & 0.465 & 0.465 & 0.363 & 0.362 & 0.434 & 0.212 & 0.430 & 0.384 & 0.103 & 0.253 \\
Gemma     & 0.109 & 0.109 & 0.115 & 0.115 & 0.451 & 0.451 & 0.102 & 0.102 & 0.499 & 0.304 & 0.255 & 0.216 & 0.398 & 0.349 \\
Falcon    & 0.492 & 0.336 & 0.523 & 0.377 & 0.524 & 0.380 & 0.307 & 0.159 & 0.304 & 0.073 & 0.430 & 0.265 & 0.219 & 0.307 \\
CodeLlama & 0.627 & 0.593 & 0.579 & 0.510 & 0.574 & 0.530 & 0.356 & 0.340 & 0.484 & 0.247 & 0.524 & 0.444 & 0.271 & 0.346 \\
\midrule
Mean  & 0.401 & 0.361 & 0.430 & 0.371 & 0.465 & 0.427 & 0.287 & 0.254 & 0.420 & 0.197 \\
Range & 0.519 & 0.484 & 0.464 & 0.395 & 0.265 & 0.228 & 0.261 & 0.258 & 0.195 & 0.231 \\
\bottomrule
\end{tabular}\vspace{-3mm}
\end{table*}

\subsection{Confusion Categories}

Each model prediction is categorized into one of several outcome types: a \textbf{True Positive (TP)} corresponds to code that is correctly classified as \vul, while a \textbf{True Negative (TN)} denotes code that is correctly classified as \safe. In contrast, a \textbf{False Positive (FP)} occurs when code is incorrectly classified as \vul, and a \textbf{False Negative (FN)} occurs when vulnerable code is incorrectly classified as \safe. We additionally distinguish \textbf{Unknown False Negatives (UnFN)}, which capture vulnerable code instances for which the model fails to produce a definitive \vul verdict under the enforced output protocol, including explicit abstentions or responses that avoid committing to a classification. Finally, \textbf{Incorrect} outputs refer to cases in which the model does not provide a definitive classification despite the enforced output protocol, such as refusals, excessive meta-reasoning without a verdict, or responses that avoid committing to either \safe or \vul; these cases reflect model behavior under the given prompt rather than parser failure. 
\textbf{UnFN} explicitly captures abstentions on vulnerable samples because abstaining on vulnerable code has similar operational consequences to false negatives in security settings.

\subsection{Metric Definitions}

Let TP, TN, FP, FN, UnFN, and Incorrect denote the outcome counts defined above. Using these quantities, we compute standard and abstention-aware evaluation metrics as follows. \textbf{Accuracy} is defined as the fraction of correctly classified instances, (TP + TN) / (TP + TN + FP + FN + UnFN). \textbf{Precision} measures the reliability of positive predictions and is computed as TP / (TP + FP), while \textbf{Recall} captures the model's ability to identify vulnerable code and is given by TP / (TP + FN + UnFN), explicitly accounting for unknown false negatives. Incorrect outputs are excluded from the accuracy denominator because they represent protocol non-compliance rather than a committed \safe/\vul classification; their operational impact is captured through abstention, coverage, and effective F1. The \textbf{F1 Score} is the harmonic mean of precision and recall, computed as 2 \(\cdot\) (Precision \(\cdot\) {Recall}) / ({Precision} + {Recall}). To quantify non-commitment behavior, we define the \textbf{Abstention Rate} as (Incorrect + UnFN) / (TP + TN + FP + FN + UnFN + Incorrect), with \textbf{Coverage} given by one minus the abstention rate. Finally, we report \textbf{Effective F1}, defined as the product of the F1 score and coverage, which jointly captures classification quality and the model's willingness to issue definitive verdicts.

\subsection{Interpretation}

Standard metrics such as accuracy and F1 alone can be misleading when models abstain frequently. A model may achieve high F1 on answered samples while remaining unusable in practice due to low coverage. The \textbf{Effective F1} metric penalizes such behavior by scaling F1 by coverage, yielding a measure that better reflects operational utility. By explicitly accounting for \texttt{UNKNOWN} outputs and abstentions, these metrics align evaluation with real-world security workflows, where failure to flag a vulnerability is often more costly than issuing a false alarm. This formulation allows \ours to distinguish between models that appear strong under closed-world assumptions and those that provide consistent, actionable outputs in realistic deployment scenarios.

\section{Supplementary Result Tables}
\label{appendix:supplementary-results}

This appendix reports numeric tables that duplicate or expand the figure-level summaries in the main text. They are included for exact lookup without requiring the main paper to repeat the same information in both graphical and tabular form.

\begin{table}[htbp]
\centering
\caption{Selected model--prompt pairs illustrating the precision--recall tradeoff.}\label{tab:tradeoff}\vspace{-3mm}
\footnotesize
\begin{tabular}{lcccccc}
\toprule
Model / Prompt & Acc. & Prec. & Recall & F1 & Eff. F1 & Abst. \\
\midrule
CodeLlama / ZS    & 0.472 & 0.492 & 0.865 & 0.627 & 0.593 & 5.47\% \\
CodeLlama / A-CoT & 0.484 & 0.492 & 0.279 & 0.356 & 0.340 & 4.48\% \\
Falcon / CoT      & 0.430 & 0.508 & 0.541 & 0.524 & 0.380 & 27.49\% \\
Falcon / S-C      & 0.202 & 0.517 & 0.216 & 0.304 & 0.073 & 75.96\% \\
Mistral / CoT     & 0.515 & 0.517 & 0.423 & 0.465 & 0.465 & 0.08\% \\
Mistral / S-C     & 0.342 & 0.514 & 0.375 & 0.434 & 0.212 & 51.07\% \\
\bottomrule
\end{tabular}\vspace{-3mm}
\end{table}

\begin{table}[htbp]
\centering
\small
\caption{Output-protocol ablation: verdict-first (VF) vs.\ verdict-last (VL) on a 1,232-sample subset. Mean rows summarize across all model--prompt combinations.}
\label{tab:ablation}\vspace{-3mm}
\footnotesize\setlength{\tabcolsep}{4pt}
\begin{tabular}{llcccccc}
\toprule
& & \multicolumn{3}{c}{\textbf{Verdict-first (VF)}} & \multicolumn{3}{c}{\textbf{Verdict-last (VL)}} \\
\textbf{Model} & \textbf{Prompt} & \textbf{F1} & \textbf{Eff.\,F1} & \textbf{Abst.} & \textbf{F1} & \textbf{Eff.\,F1} & \textbf{Abst.} \\
\midrule
CodeLlama & ZS & 0.616 & 0.571 & 7.2 & 0.538 & 0.460 & 14.5 \\
          & FS  & 0.307 & 0.288 & 6.2 & 0.511 & 0.464 & 9.3 \\
          & CoT       & 0.605 & 0.565 & 6.7 & 0.562 & 0.424 & 24.7 \\
          & A-CoT     & 0.633 & 0.592 & 6.3 & 0.448 & 0.314 & 30.0 \\
          & S-C       & 0.631 & 0.592 & 6.2 & 0.448 & 0.309 & 31.0 \\
\midrule
Mistral   & ZS & 0.580 & 0.579 & 0.2 & 0.364 & 0.204 & 44.0 \\
          & FS  & 0.424 & 0.382 & 10.0 & 0.166 & 0.087 & 47.2 \\
          & CoT       & 0.575 & 0.575 & 0.2 & 0.434 & 0.217 & 50.0 \\
          & A-CoT     & 0.223 & 0.222 & 0.2 & 0.270 & 0.106 & 60.9 \\
          & S-C       & 0.232 & 0.232 & 0.2 & 0.248 & 0.093 & 62.3 \\
\midrule
DeepSeek  & ZS & 0.083 & 0.080 & 3.2 & 0.413 & 0.346 & 16.3 \\
          & FS  & 0.551 & 0.500 & 9.2 & 0.408 & 0.302 & 26.0 \\
          & CoT       & 0.448 & 0.445 & 0.6 & 0.433 & 0.323 & 25.4 \\
          & A-CoT     & 0.459 & 0.459 & 0.0 & 0.304 & 0.268 & 11.8 \\
          & S-C       & 0.441 & 0.441 & 0.0 & 0.306 & 0.268 & 12.3 \\
\midrule
Gemma     & ZS & 0.144 & 0.144 & 0.0 & 0.374 & 0.249 & 33.6 \\
          & FS  & 0.019 & 0.019 & 0.0 & 0.139 & 0.097 & 29.8 \\
          & CoT       & 0.523 & 0.523 & 0.0 & 0.507 & 0.392 & 22.6 \\
          & A-CoT     & 0.031 & 0.031 & 0.2 & 0.108 & 0.074 & 31.1 \\
          & S-C       & 0.034 & 0.034 & 0.1 & 0.102 & 0.071 & 30.0 \\
\midrule
Falcon    & ZS & 0.410 & 0.249 & 39.3 & 0.411 & 0.236 & 42.6 \\
          & FS  & 0.455 & 0.307 & 32.6 & 0.432 & 0.263 & 39.0 \\
          & CoT       & 0.461 & 0.291 & 36.8 & 0.358 & 0.198 & 44.7 \\
          & A-CoT     & 0.413 & 0.280 & 32.1 & 0.343 & 0.214 & 37.6 \\
          & S-C       & 0.412 & 0.276 & 33.0 & 0.339 & 0.210 & 38.1 \\
\midrule
\textbf{Mean} &  & \textbf{0.403} & \textbf{0.347} & \textbf{9.2} & \textbf{0.344} & \textbf{0.248} & \textbf{32.6} \\
\bottomrule
\end{tabular}\vspace{-3mm}
\end{table}

\subsection{Additional Raw Results}
\label{appendix:graphs-tables}

\begin{table}[t]
\captionsetup{justification=centering}
\caption{Extended confusion statistics by model and prompt template.}
\label{tab:appendix-confusion}\vspace{-3mm}
\centering
\scalebox{0.7}{\begin{tabular}{
  >{\raggedright\arraybackslash}l
  >{\raggedright\arraybackslash}p{1.6cm}
  c c c c c c
}
\hline
\textbf{Model} & \textbf{Prompt} & \textbf{TP} & \textbf{TN} & \textbf{FP} & \textbf{FN} & \textbf{Incorrect} & \textbf{UnFN} \\
\hline

\multirow{5}{*}{DeepSeek}
& ZS           & 679  & 2307 & 686  & 2326 & 44   & 32   \\
& FS            & 1508 & 1158 & 1401 & 1137 & 478  & 392  \\
& CoT     & 690  & 2286 & 725  & 2305 & 26   & 42   \\
& A-CoT         & 686  & 2300 & 722  & 2333 & 15   & 18   \\
& S-C     & 901  & 344  & 796  & 330  & 1897 & 1806 \\
\hline

\multirow{5}{*}{Mistral}
& ZS            & 1331 & 1677 & 1339 & 1684 & 21   & 22   \\
& FS             & 1091 & 1973 & 1020 & 1898 & 44   & 48   \\
& CoT     & 1280 & 1840 & 1195 & 1745 & 2    & 3    \\
& A-CoT         & 849  & 2238 & 792  & 2180 & 7    & 8    \\
& S-C     & 1138 & 409  & 1075 & 350  & 1553 & 1549 \\
\hline

\multirow{5}{*}{Gemma}
& ZS            & 183  & 2884 & 153  & 2854 & 0    & 0    \\
& FS             & 192  & 2932 & 105  & 2845 & 0    & 0    \\
& CoT     & 1275 & 1701 & 1336 & 1762 & 0    & 0    \\
& A-CoT         & 172  & 2860 & 177  & 2865 & 0    & 0    \\
& S-C     & 1476 & 452  & 1399 & 369  & 1186 & 1192 \\
\hline

\multirow{5}{*}{Falcon}
& ZS           & 1459 & 609  & 1433 & 650  & 995  & 928  \\
& FS             & 1641 & 582  & 1597 & 554  & 858  & 842  \\
& CoT     & 1642 & 611  & 1591 & 560  & 835  & 835  \\
& A-CoT         & 677  & 849  & 700  & 921  & 1488 & 1439 \\
& S-C     & 655  & 102  & 611  & 92   & 2324 & 2290 \\
\hline

\multirow{5}{*}{CodeLlama}
& ZS            & 2626 & 166  & 2711 & 239  & 160  & 172  \\
& FS             & 2139 & 457  & 2210 & 545  & 370  & 353  \\
& CoT     & 1955 & 787  & 1996 & 686  & 254  & 214  \\
& A-CoT         & 848  & 2028 & 874  & 2053 & 136  & 136  \\
& S-C     & 1445 & 77   & 1495 & 81   & 1465 & 1511 \\
\hline
\end{tabular}}\vspace{-2mm}
\end{table}

\begin{table}[htbp]
\centering
\caption{Classification performance by model and prompt.}
\label{tab:appendix-metrics}\vspace{-3mm}
\scalebox{0.7}{\begin{tabular}{l l c c c}
\hline
\textbf{Model} & \textbf{Prompt} & \textbf{Accuracy} & \textbf{Precision} & \textbf{Recall} \\
\hline

\multirow{5}{*}{DeepSeek}
& Zero-shot        & 0.495 & 0.497 & 0.224 \\
& Few-shot         & 0.476 & 0.518 & 0.497 \\
& CoT & 0.492 & 0.488 & 0.227 \\
& A-CoT     & 0.493 & 0.487 & 0.226 \\
& S-C & 0.298 & 0.531 & 0.297 \\
\hline

\multirow{5}{*}{Mistral}
& Zero-shot        & 0.497 & 0.499 & 0.438 \\
& Few-shot         & 0.508 & 0.517 & 0.359 \\
& CoT & 0.515 & 0.517 & 0.423 \\
& A-CoT     & 0.509 & 0.517 & 0.280 \\
& S-C & 0.342 & 0.514 & 0.375 \\
\hline

\multirow{5}{*}{Gemma}
& Zero-shot        & 0.505 & 0.545 & 0.060 \\
& Few-shot         & 0.514 & 0.646 & 0.063 \\
& CoT & 0.490 & 0.488 & 0.420 \\
& A-CoT     & 0.499 & 0.493 & 0.057 \\
& S-C & 0.394 & 0.513 & 0.486 \\
\hline

\multirow{5}{*}{Falcon}
& Zero-shot        & 0.407 & 0.504 & 0.480 \\
& Few-shot         & 0.426 & 0.507 & 0.540 \\
& CoT & 0.430 & 0.508 & 0.541 \\
& A-CoT     & 0.333 & 0.492 & 0.223 \\
& S-C & 0.202 & 0.517 & 0.216 \\
\hline

\multirow{5}{*}{CodeLlama}
& Zero-shot        & 0.472 & 0.492 & 0.865 \\
& Few-shot         & 0.455 & 0.492 & 0.704 \\
& CoT & 0.486 & 0.495 & 0.685 \\
& A-CoT     & 0.484 & 0.492 & 0.279 \\
& S-C & 0.330 & 0.491 & 0.476 \\
\hline
\end{tabular}}\vspace{-5mm}
\end{table}

Table~\ref{tab:appendix-confusion} summarizes true positive (TP), true negative (TN), false positive (FP),
false negative (FN), Incorrect, and UnFN outcome distributions across all evaluated configurations,
enabling fine-grained analysis of detection reliability and model decision behavior. Self-consistency
prompting produces the largest increases in unresolved and error-prone outputs, reflected by elevated
Incorrect and UnFN counts across multiple models, indicating reduced vulnerability recall and increased
prediction instability. In contrast, zero-shot and Chain-of-Thought prompting maintain more balanced
outcome distributions for several model families, preserving higher volumes of correct classifications
with fewer unresolved predictions. These outcome shifts align with the Effective F1 and coverage trends
reported in the main analysis, as growth in Incorrect and UnFN outcomes reduces usable prediction volume
and contributes to the performance degradation observed under fragile prompting strategies.

Table~\ref{tab:appendix-metrics} summarizes performance across all evaluated model--prompt combinations
using accuracy, precision, and recall. The results show that prompting strategy systematically shapes
performance patterns across models. In particular, few-shot prompting generally increases recall compared
to zero-shot settings, indicating improved detection coverage. Chain-of-Thought prompting tends to produce
more balanced outcomes, stabilizing accuracy while moderating fluctuations in recall. Adaptive
Chain-of-Thought has mixed effects: it preserves accuracy in some configurations but leads to notable
recall degradation in others. Self-consistency introduces a distinct trade-off, often reducing accuracy
while maintaining relatively stable precision---consistent with the more conservative prediction behavior
observed in the abstention analysis.

From a metric perspective, overall accuracy remains relatively stable across strategies (typically
0.47--0.52 outside self-consistency), suggesting limited sensitivity of aggregate correctness to prompt
structure. Precision shows moderate variability, generally remaining near 0.50, indicating robust control
of false positives. In contrast, recall exhibits the largest variation, ranging from 0.057 (Gemma Adaptive
CoT) to 0.865 (CodeLlama Zero-Shot), highlighting detection coverage as the main driver of performance
variability.

Table~\ref{tab:abstention-rate} reports abstention behavior across all evaluated model--prompt
combinations and directly reflects effective prediction coverage, where lower abstention corresponds to higher usable output coverage. Our A-CoT-based self-consistency configuration produces the highest abstention rates for every model, exceeding 39\% in all cases and reaching 75.96\% for Falcon, indicating a substantial reduction in effective coverage under this majority-vote policy. In contrast, the evaluated zero-shot and Chain-of-Thought templates maintain low abstention for models such as Mistral and Gemma, often below 1\%, resulting in near-complete output coverage. Falcon exhibits the highest mean abstention rate (42.26\%), reflecting reduced prediction availability across template choices, whereas Gemma preserves near-full coverage under most evaluated prompt templates. These coverage reductions directly contribute to the Effective F1 trends reported in the main results, as templates or aggregation policies that increase abstention reduce the proportion of evaluable predictions and therefore lower the effective performance even when raw F1 remains stable.

\end{document}